\documentclass{article}

\usepackage[preprint]{neurips_2025}

\usepackage[utf8]{inputenc} 
\usepackage[T1]{fontenc}    
\usepackage{hyperref}       
\usepackage{url}            
\usepackage{booktabs}       
\usepackage{amsfonts}       
\usepackage{nicefrac}       
\usepackage{microtype}      
\usepackage{xcolor}         
\usepackage{graphicx}
\usepackage{natbib}
\usepackage{cleveref}       
\usepackage{multirow}
\usepackage{soul}
\usepackage{wrapfig}
\usepackage{subcaption}
\usepackage{wrapfig}
\usepackage{textcmds}

\crefname{figure}{Fig.}{Figs.}
\crefname{table}{Tab.}{Tabs.}

\title{Generalist Multi-Class Anomaly Detection via Distillation to Two Heterogeneous Student Networks}

\author{%
  Hangil Park \quad Yongmin Seo \quad Tae-Kyun Kim \\
    School of Computing, KAIST \\
    \texttt{\{hangil.park, yongmin.seo, kimtaekyun\}@kaist.ac.kr}
}

\begin{document}

\maketitle

\begin{abstract}
     Anomaly detection (AD) plays an important role in various real-world applications. Recent advancements in AD, however, are often biased towards industrial inspection, struggle to generalize to broader tasks like semantic anomaly detection and vice versa. Although recent methods have attempted to address general anomaly detection, their performance remains sensitive to dataset-specific settings and single-class tasks. In this paper, we propose a novel dual-model ensemble approach based on knowledge distillation (KD) to bridge this gap. Our framework consists of a teacher and two student models: an Encoder-Decoder model, specialized in detecting patch-level minor defects for industrial AD and an Encoder-Encoder model, optimized for semantic AD. Both models leverage a shared pre-trained encoder (DINOv2) to extract high-quality feature representations. The dual models are jointly learned using the Noisy-OR objective, and the final anomaly score is obtained using the joint probability via local and semantic anomaly scores derived from the respective models. We evaluate our method on eight public benchmarks under both single-class and multi-class settings: MVTec-AD, MVTec-LOCO, VisA and Real-IAD for industrial inspection and CIFAR-10/100, FMNIST and View for semantic anomaly detection. The proposed method achieved state-of-the-art accuracies in both domains, in multi-class as well as single-class settings, demonstrating generalization across multiple domains of anomaly detection. Our model achieved an image-level AUROC of 99.7\% on MVTec-AD and 97.8\% on CIFAR-10, which is significantly better than the prior general AD models in multi-class settings and even higher than the best specialist models on individual benchmarks.
\end{abstract}

\section{Introduction}
Anomaly Detection (AD), is a critical task across a wide range of applications, from industrial quality control to cybersecurity, healthcare, and autonomous systems. The goal of AD is to identify patterns that deviate from the norm, representing potential defects, attacks, or abnormalities. As such, detecting anomalies can prevent system failures, improve security, and ensure quality and reliability in automated processes. One area where AD has been particularly impactful is industrial inspection, where AD systems play a key role in automating the detection of manufacturing defects. In these contexts, anomalies such as scratches, dents, or missing components need to be identified with a high precision to ensure product quality and reduce costs. Despite the success of AD methods in such specialized tasks, a major challenge remains: generalizability across domains.

\begin{figure}[h]
    \begin{minipage}{0.25\linewidth}
  \centering
  \includegraphics[width=0.9\linewidth]{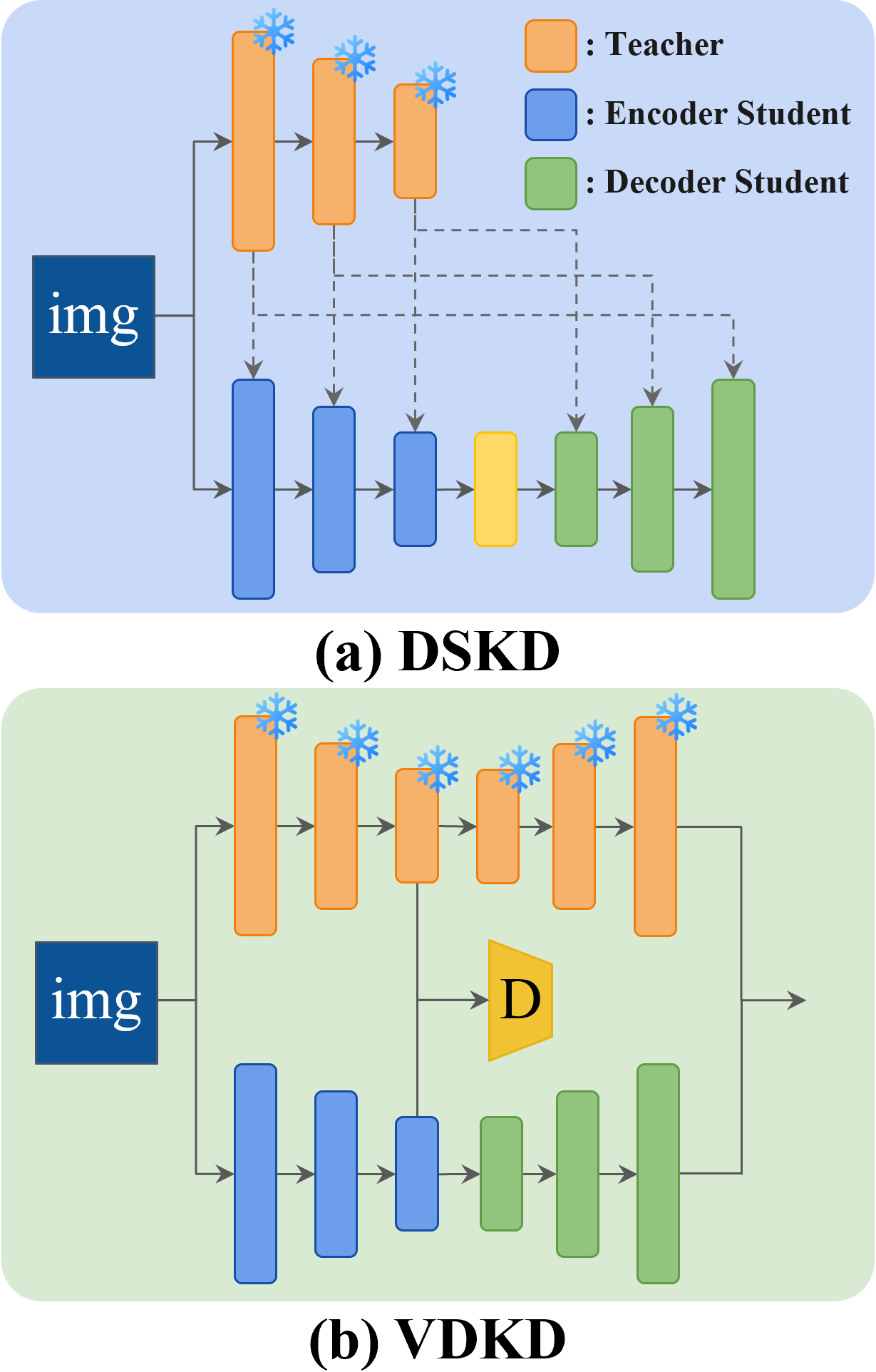}
  \vspace{-1mm}
   \caption{\textbf{Prior KD methods with dual students.} (a) DSKD \citep{dualstudent}, (b) VDKD \citep{VDKD}.}
   \vspace{-5mm}
  \label{fig:figure1} 
    \end{minipage}
    \hfill
    \begin{minipage}{0.72\linewidth}
    \begin{center}
    \includegraphics[width=\linewidth]{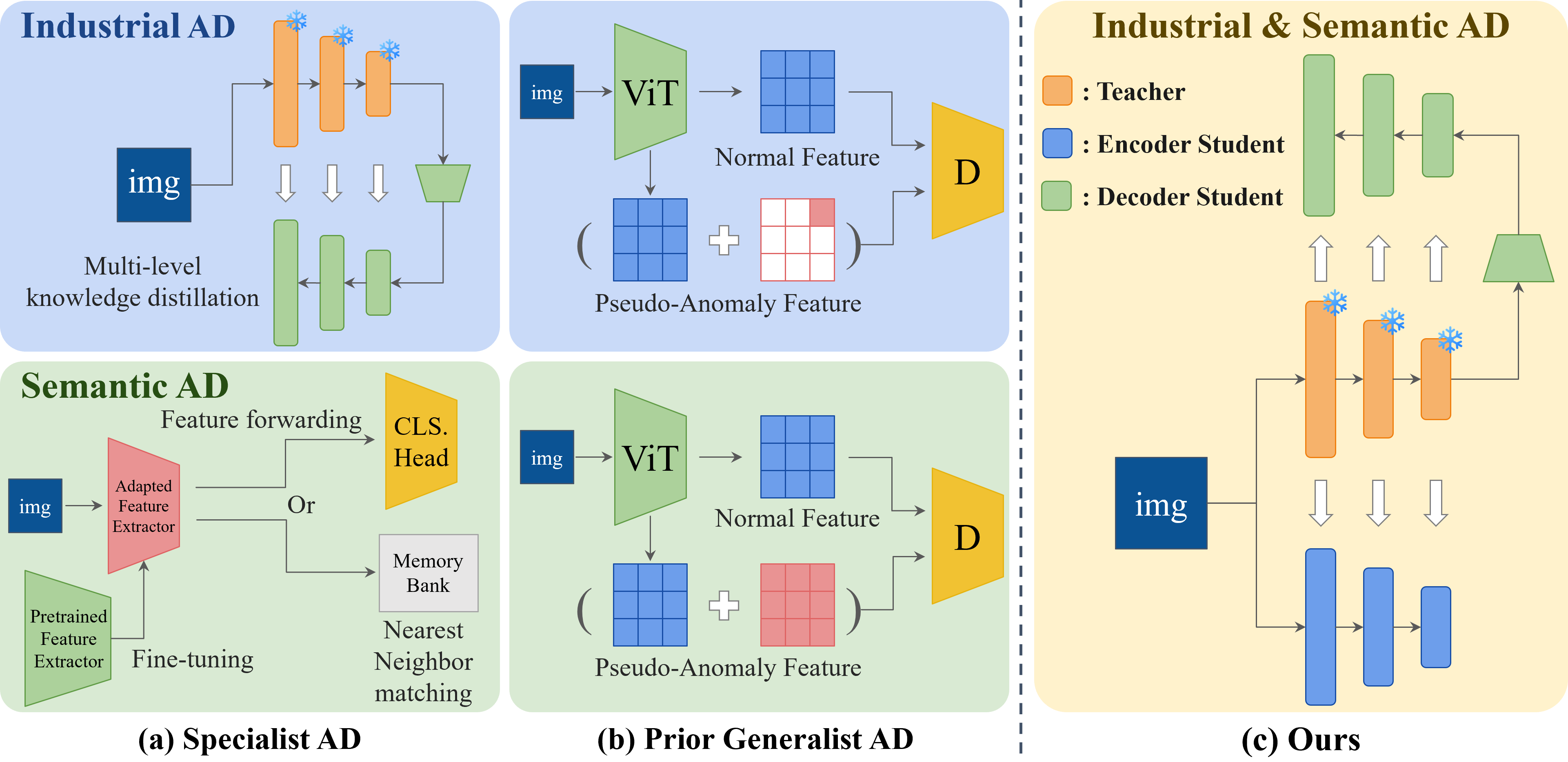}
    \end{center}
    \vspace{-3mm}
    \caption{\textbf{Comparison among previous anomaly detection methods and ours.} (a) Specialist AD: KD methods(top), major semantic AD methods(bottom), (b) Prior Generalist AD \citep{generalad}, (c) Proposed Generalist AD.}
    \vspace{-5mm}
    \label{fig:figure0}
    \end{minipage}
\end{figure}

A major group of recent work in anomaly detection, particularly those based on datasets like MVTec-AD \citep{mvtecad}, have concentrated on industrial applications. These methods commonly employ techniques such as patch embedding \citep{patchcore, simplenet, spade, padim}, pseudo-anomaly generation \citep{cutpaste, fitymi,nsa, dream, realnet} and knowledge distillation (KD) \citep{mkd, rd4ad, mva, rd4ad++, dinomaly, destseg, liu2024dual, zhang2024contextual, uninformed, dualstudent, VDKD, EfficientAD}. Recent KD-based methods \citep{rd4ad, rd4ad++, dinomaly} show state-of-the-arts performance on industrial anomaly detection (AD). In these methods, input samples are processed through a pre-trained teacher model, a bottleneck, and a decoder student model. The bottleneck and student decoder model are trained to reconstruct the teacher's multi-level features \citep{rd4ad}. The KD methods \citep{dualstudent, EfficientAD, VDKD} also adopted dual-student models shown in \cref{fig:figure1}. These methods employ two student models to enhance representation learning. Specifically, they leverage these models for extracting richer feature representations \citep{dualstudent}, guiding the student to capture global contextual information through an autoencoder, and learning discriminative features using MAE \citep{VDKD}. While the aforementioned methods yield impressive accuracies in detecting specific defect types for industrial AD, they often underperform in broader anomaly detection tasks involving multiple classes or in domains that require semantic understanding of data e.g., CIFAR-10/100.

Recently, a few approaches have aimed to achieve state-of-the-art performance in anomaly detection across both industrial inspection and semantic anomaly datasets \citep{uniformaly, generalad}. These methods, however, depend on specific parameter choices, backbone networks, or dataset-specific prior knowledge. GeneralAD \citep{generalad} achieves general anomaly detection via the ViT feature extractor and patch-wise discriminator. Performance of GeneralAD heavily relies on synthetic anomaly generation (See the \cref{sec:sup_ablation_generalAD} for details). They use different anomaly generation strategies for different domains (e.g., add noise to a random patch feature for industrial AD and add noise to all patches for semantic AD). Additionally, the previous methods primarily focus on one-vs-rest setting, rather than more challenging and generalized multi-class settings.

In this paper, we propose a novel knowledge distillation-based approach that addresses the gap from existing industrial anomaly detection to general multi-class anomaly detection.
It exploits an encoder-decoder model for detecting local defects in industrial settings and an encoder-encoder model for semantic anomaly detection, preserving the semantic integrity of input images, as shown in \cref{fig:figure0}(c). This dual-model framework achieves state-of-the-art performance on both industrial and semantic anomaly detection tasks, in multi-class as well as single-class settings, demonstrating the generalizability and robustness of our method across diverse domains. The main contributions are as follows:
\begin{itemize} 
    \item We propose a novel dual KD-based architecture and learning method for generalist anomaly detection,  capable of handling a variety of anomaly types.
    \item In the encoder-encoder pair, we introduce the class token discrepancy for better detecting semantic-level 
    irregularities. 
    The encoder architecture is advantageous to represent the whole image semantic contexts. 
    \item The decoder student which maps back to image pixels convey local information, thus being better at detecting local defects in industrial settings. 
    \item The proposed method is significantly better than previous general AD methods in multi-class settings, and on par with the best specialist models on each benchmark. Our method achieves SOTA results on six out of eight public benchmarks.
    \item Our method also demonstrates sample efficiency when only few-shots are available per class. 
\end{itemize}

\section{Related work}
\label{sec:related_work}

\noindent \textbf{Industrial Defect Detection.} In these applications, anomalies include manufacturing defects such as cracks, scratches, and surface irregularities, which need to be detected with precision to ensure product quality. Recent methods mainly rely on unsupervised learning, where models are trained using only normal data. These methods commonly employ techniques such as patch embedding \cite{patchcore, simplenet, spade, padim}, pseudo-anomaly generation \cite{cutpaste, fitymi,nsa, dream, realnet, simplenet}, contrastive learning \cite{rd4ad++, recontrast} and knowledge distillation (KD) \cite{mkd, rd4ad, rd4ad++, dinomaly, destseg, liu2024dual, zhang2024contextual, uninformed, dualstudent, EfficientAD, VDKD}. In the KD based methods, the student model learns to imitate the teacher's feature and is expected to reconstruct well normal samples and struggle for abnormal samples. MKD \cite{mkd} first suggests KD based anomaly detection by adopting a shallow student encoder model in the distillation framework. RD4AD \cite{rd4ad} shows that the reverse distillation framework using a bottleneck and deep decoder student model is a better option for industrial defect detection. UniAD \cite{uniad} proposed a reconstruction-based multi-class method via a customized ViT architecture. 
UniAD addresses "identical shortcut" issues in industrial AD, where models tend to learn an identity mapping rather than reconstruction of normal samples. The shortcut reconstructs well for anomalous samples either, leading to a low detection accuracy. 
Dinomaly \cite{dinomaly} also falls to the KD category and and raises concerns that dense, layer-to-layer distillation may cause the student model to overly imitate the teacher’s behavior, potentially exacerbating identical shortcut problems, as noted in \cite{liang2024module}. Instead, it advocates a looser loss formulation for KD. \cite{dualstudent, EfficientAD, VDKD} utilize two different types of student models with knowledge distillation.
\cite{dualstudent} employs an encoder student and a decoder student, with the teacher distilling knowledge to both. While the encoder student receives the same input as the teacher network, the decoder student takes the output of the encoder student as its input.
\cite{EfficientAD} consists of three components: a teacher, an autoencoder, and a student. The teacher and autoencoder distill local and global information to the student, respectively.
\cite{VDKD} is based on the ViT-MAE architecture, where the teacher distills knowledge to the student at both the encoder and decoder levels. For the encoder output, the student is guided by a discriminator that determines whether the feature originates from the teacher, whereas for the decoder output, it is guided using pixel-level labels (e.g., anomaly maps). All methods are, however, tested on industrial AD. In contrast, our proposed method employs two explicit student networks that complement each other, and is demonstrated for semantic as well as industrial AD. Several methods \cite{uniad, mint-ad, moead} use multi-class settings where a single AD model is trained over all classes of a benchmark than a model per class. \cite{moead} applies Mixture-of-Experts framework on intermidiate layers to deal with various input types of multi-class industrial AD. They are yet less applicable to semantic AD tasks where more comprehensive understanding of images is required.

\noindent\textbf{Semantic Anomaly Detection.} Semantic AD, often framed as one-class classification problems or Out-of-Distribution detection, tackles cases where anomalies look normal alone but abnormal and normal samples belong to different semantic classes. For example, in CIFAR-10, the class of dog can be defined abnormal against other normal classes e.g., cat. 

Unlike the industrial inspection, 

semantic AD requires understanding of contextual relations among objects and surroundings.
Transformaly \cite{transformaly} achieves SOTA performance on CIFAR-10 by leveraging two feature spaces: one extracted from a pre-trained Vision Transformer (ViT) and another derived from a teacher-student framework. Additionally, \cite{msc} introduces the mean-shifted contrastive loss specifically designed for anomaly detection, while FITYMI \cite{fitymi} employs diffusion-based data augmentation to generate pseudo-anomalies for supervised learning. Despite their success in semantic anomaly detection, these methods perform poorly in industrial settings, where normal and abnormal samples often share common semantic features. 

In the survey \cite{openood, openood1.5} several works \cite{mls_klm, knn, vim, dice} are categorised to post-hoc methods that achieve SOTA semantic AD accuracies exploiting the powerful backbone networks.

\noindent\textbf{General Anomaly Detection}. Recently, a few methods have been proposed for both industrial inspection and semantic anomaly detection \cite{uniformaly, generalad}. Uniformaly \cite{uniformaly} enhances anomaly detection by eliminating less informative background patches through Back Patch Masking (BPM), followed by anomaly identification using a memory bank based top k-ratio feature matching technique. GeneralAD \cite{generalad} employs a patch-wise discriminator, which classifies input patch features as normal or anomalous. To train the discriminator, normal and abnormal patch features are required. The normal patch features are extracted from normal samples using a pre-trained ViT, while the pseudo-anomalies are generated by applying feature distortion to these normal patches.
Both methods, however, have limited adaptation to new datasets. 

Uniformaly relies heavily on explicit parameters (e.g., k-ratio), and its optimal backbone network varies depending on datasets. GeneralAD's pseudo-anomaly generation strategy is contingent on dataset prior knowledge. In contrast, our method operates without reliance on dataset-specific prior knowledge and is evaluated at more challenging and general multi-class settings. 

\noindent\textbf{Adaptation}. Also, there are some few-shot methods \cite{anomalyclip, adaclip, inctrl} leveraging pre-trained Visual-Language Models (VLMs) like CLIP \cite{clip}, which demonstrate superior generalization (via adaptation) capabilities. 
These methods require manual prompts for each dataset and heavily rely on VLMs. AnomalyCLIP \cite{anomalyclip}, AdaCLIP \cite{adaclip} require auxiliary datasets containing numerous both normal and anomalous images for training and their application is limited to industrial and medical domains. In contrast, we address a novel model architecture and its training strategies that can be applied to any domains, not efficient adaptation techniques. 

\noindent\textbf{Relation to Broader Areas}. AD also relates to other areas, such as active learning \cite{al1, al2} and semi-supervised learning \cite{ssl1, ssl2}. The core idea of active learning is to select unlabeled samples that maximally improves the model if labeled. Active learning (AL) often picks them up using diversity- or outlier-based strategies and anomalies in AD appear as outliers/different points from normal data. Semi-supervised learner makes use of a large pool of unlabeled data with the small labeled set, improving generalisation by learning the structure of unlabeled data, which also grounds AD from normal data. Often proposed frameworks for SSL are adapted to AL, and vise-versa. 

\section{KD based generalist multi-class AD}
\label{sec:method}

\begin{figure*}[ht]
\begin{center}
\includegraphics[width=0.9\textwidth]{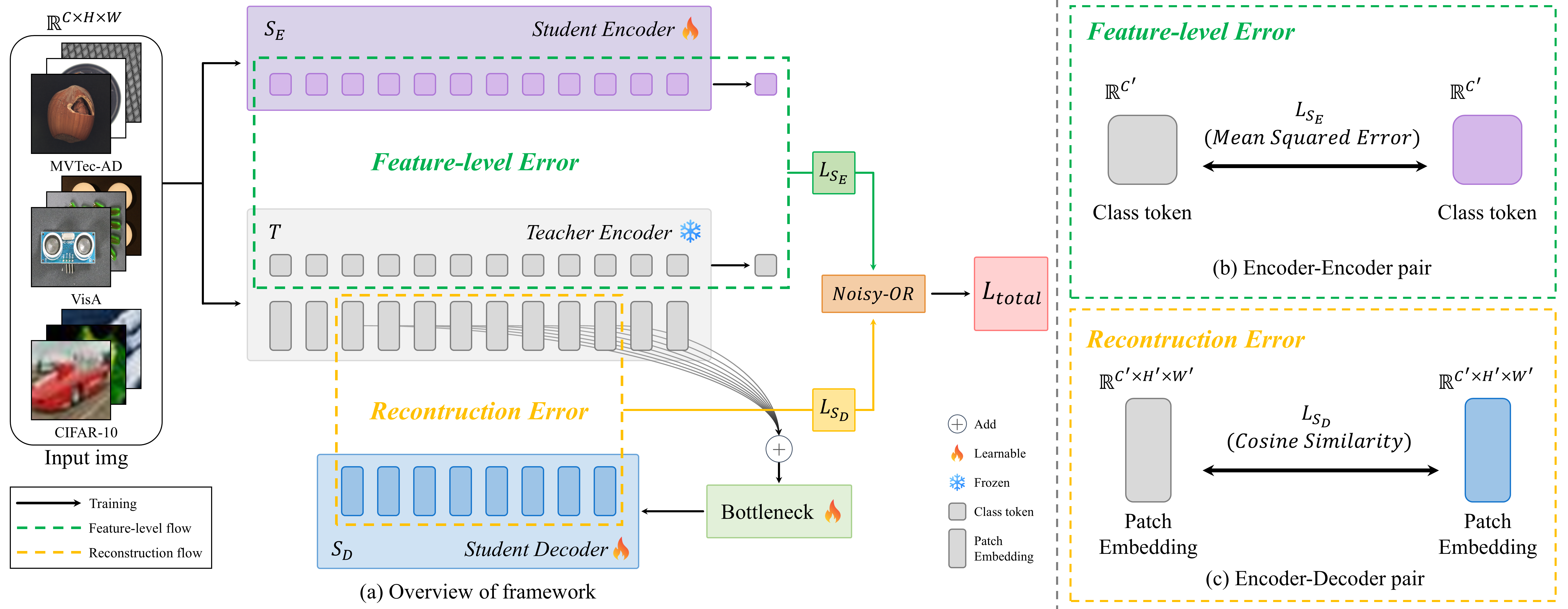}
\end{center}
\vspace*{-2mm}
\caption{\textbf{Framework overview.} Our method consists of three components: a pre-trained teacher encoder, an encoder student, and a decoder student, which form encoder-encoder and encoder-decoder pairs. 
The encoder-decoder pair is intended to detect local anomalies through the reconstruction error, while the encoder-encoder pair is designed to capture semantic anomalies by leveraging feature-level errors and class tokens. Noisy-OR function is applied to fuse the errors from each pair, enabling an unified anomaly detection framework.}
\vspace*{-5mm}
\label{fig:framework}
\end{figure*}

\subsection{Method overview}

As shown in \cref{fig:framework}, our approach combines two teacher-student models based on knowledge distillation (KD) within an ensemble framework. Specifically, we employ an Encoder-Decoder model and an Encoder-Encoder model, both of which share a pre-trained backbone for feature extraction. 
The encoder student observes the whole input images and captures semantic distributions, while the decoder student reconstructs image pixels or such representations, thus maintaining minor local changes. 

Our encoder-decoder pair model is optimized for industrial anomaly detection. This model uses an encoder to extract features from input images and a decoder to reconstruct the encoder features. Anomalies are detected based on discrepancies between the teacher's encoding and the student’s reconstruction. This model is effective in identifying structural anomalies, such as defects in industrial products, where patch-level differences are critical. 
Our encoder-encoder pair model is designed for semantic anomaly detection. This model is based on an encoder architecture for both the teacher and student networks. The teacher model is pre-trained on ImageNet using DINOv2 \cite{DINOv2, dinoreg}, while the student model is trained from scratch via distillation on target datasets. This model is capable of identifying anomalies that involve semantic irregularities by leveraging high-level feature representation and class tokenization. The two models are integrated into a shared backbone architecture that provides general feature representations, ensuring both models benefit from high-quality feature embeddings. The models are jointly trained using the Noisy-OR objective.

\subsection{Encoder-Decoder pair} The Encoder-Decoder model is tailored for industrial anomaly detection, where anomalies typically involve local structural defects that can be detected through patch-level analysis. For the patch-level analysis, we use a ViT architecture for both the teacher and student model. 

\begin{figure}[h]
    \begin{minipage}{0.34\linewidth}
    \begin{subfigure}{\linewidth}
    \centering
    \includegraphics[width=0.9\linewidth]{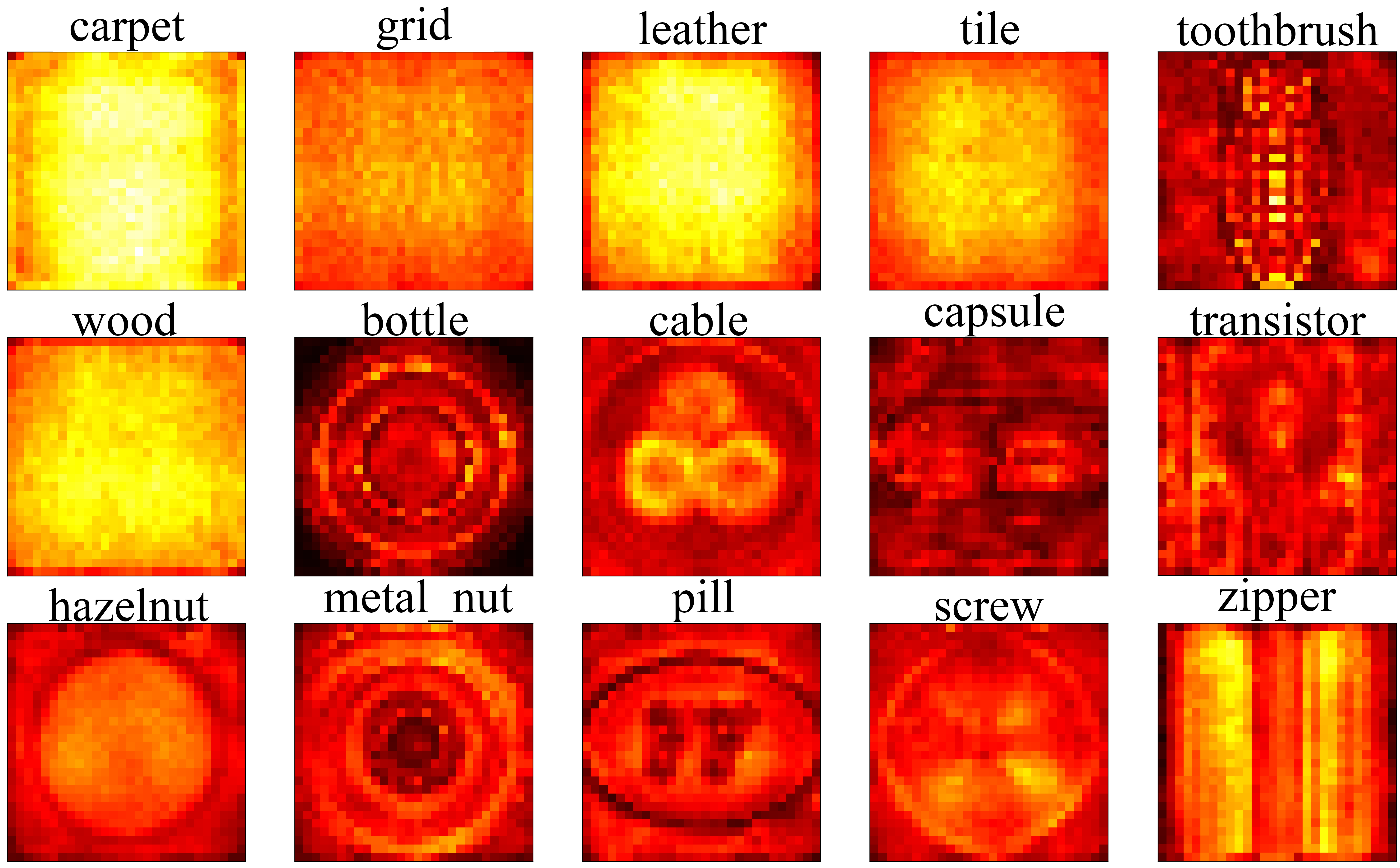}
    \vspace{-2mm}
    \caption{MVTec-AD}
    \label{fig:mvtec}
   \end{subfigure}
   \vfill
   \begin{subfigure}{\linewidth}
   \centering
    \includegraphics[width=0.9\linewidth]{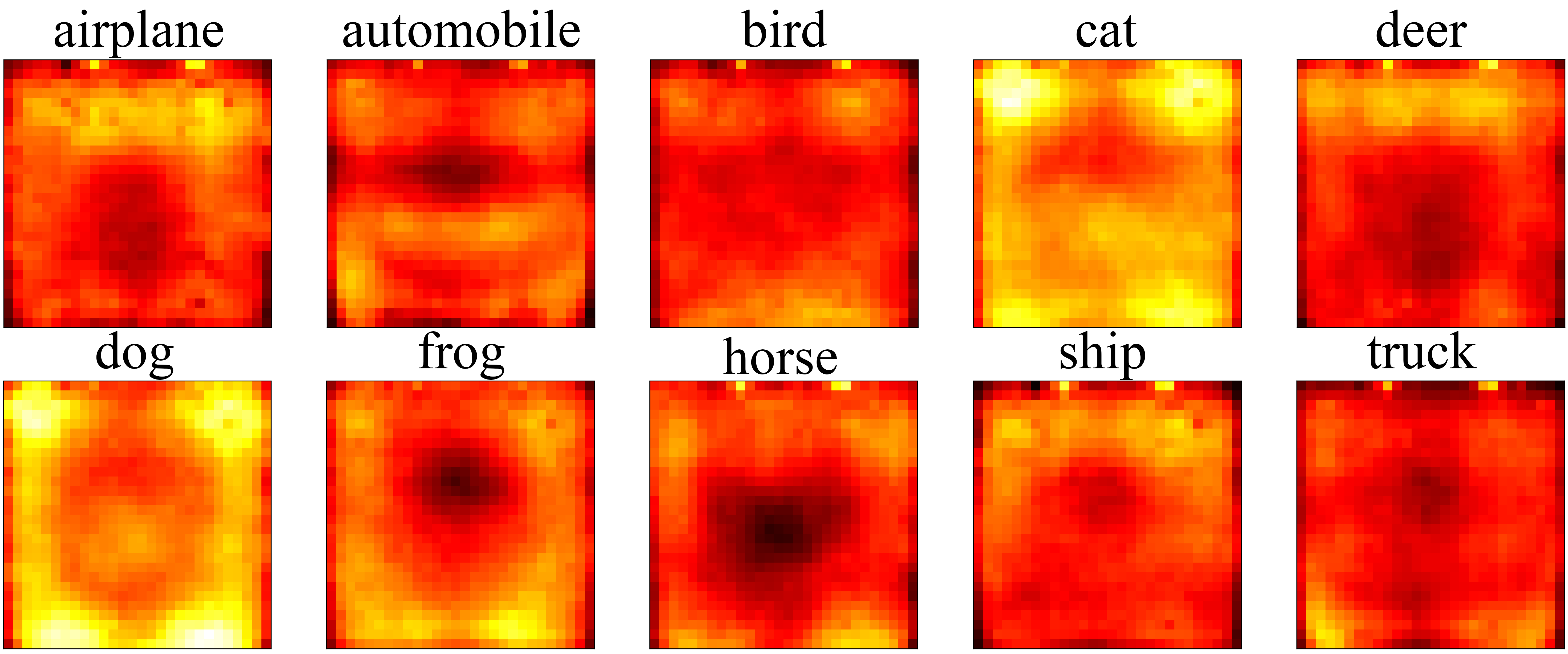}
    \vspace{-2mm}
    \caption{CIFAR-10}
    \label{fig:cifar10}
  \vspace{-2mm}
   \end{subfigure}
     \caption{\textbf{Variance of feature maps from DINOv2.} Brighter regions indicate larger variances. Normal samples from the training set are used.}
  \label{fig:figure_4_variance_of_feature_map}
  \vspace{-3mm}
  \end{minipage}
  \hfill
    \begin{minipage}{0.63\linewidth}
\begin{center}
\includegraphics[width=\linewidth]{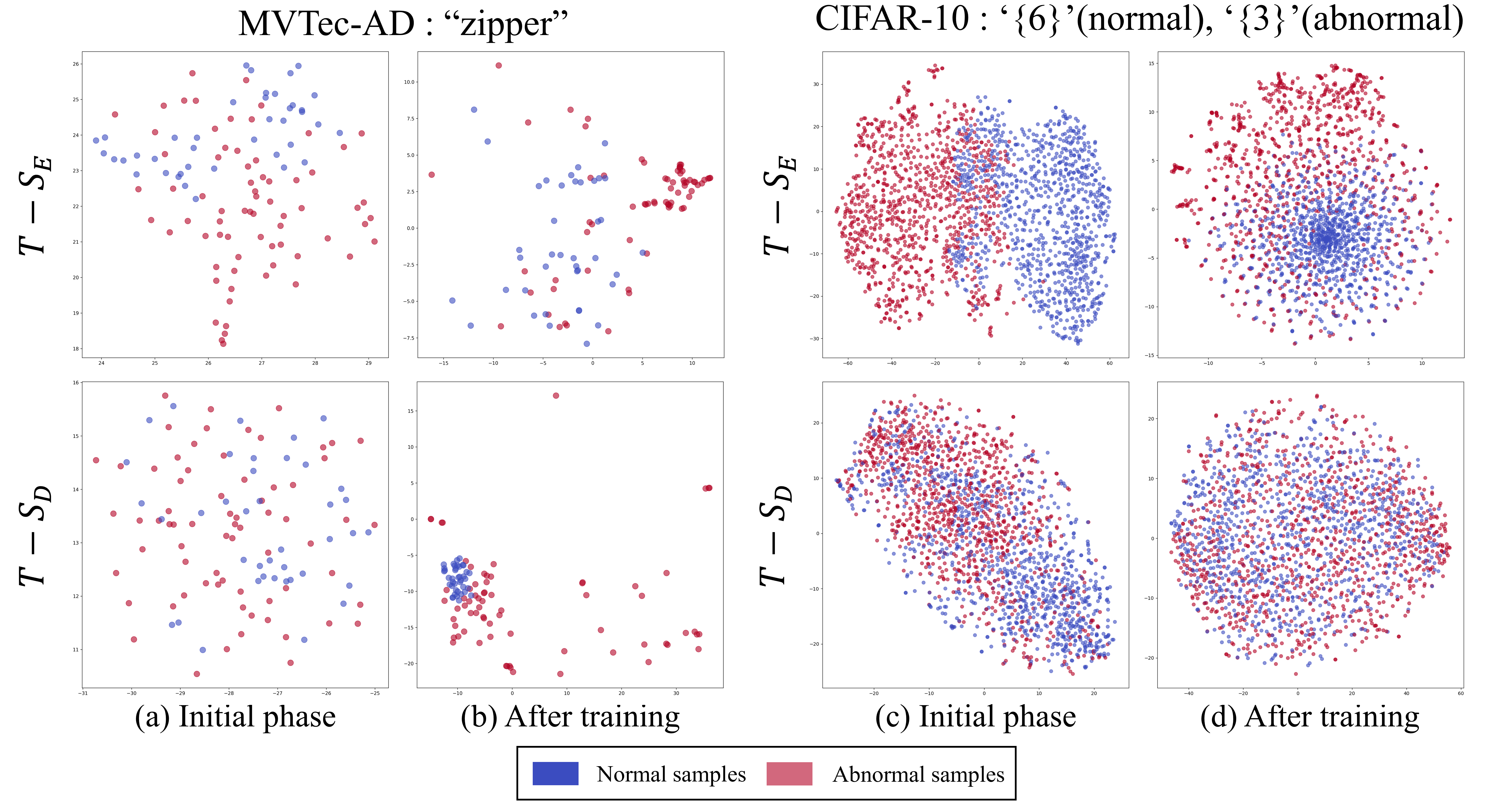}
\end{center}
\vspace*{-3mm}
\caption{\textbf{Feature t-SNE visualization.} We visualize the feature distribution shift during the training phase for MVTec-AD ((a), (b)) and CIFAR-10 ((c), (d)). $T$-$S_{E}$ and $T$-$S_{D}$ represent the feature difference between the teacher and the encoder student, and the feature difference between the teacher and the decoder student, respectively.}
\label{fig:tsne}
\vspace{-5mm}
      \end{minipage}
\end{figure}

Dinomaly \cite{dinomaly} is adopted for the student model.
The teacher model $T$ is an encoder that extracts feature embeddings from input images, while the student model $S_{D}$ is a decoder that reconstructs the input images from the teacher’s feature embeddings. The reconstruction error between the original input and the reconstructed image is used to detect anomalies. The model analyzes anomalies at the patch level, focusing on discrepancies in small, localized regions of images. 
Large reconstruction errors in specific patches indicate the presence of an anomaly in local areas. \cref{fig:figure_4_variance_of_feature_map} visualises the variance of feature embedding maps from DINOv2. We observe that the variance in feature maps for MVTec-AD dataset displays class specific patterns, whereas the feature maps for CIFAR-10 exhibit no discernible pattern. This supports the effectiveness of our decoder model and its patch-wise features, which is particularly beneficial to industrial AD scenarios. We adopt the loss function from \cite{dinomaly}, which relaxes the element-wise comparison.
This loss takes a large value when the cosine similarity between the features of teacher student is low. It forces the student model to reconstruct well the teacher features for normal input samples in the training set. The loss function for our decoder student $S_{D}$ is following:
\begin{equation}
    L_{S_{D}}(x) = \frac{1}{2}\sum_{i=1}^{2}1-\cos(vec(F_{T}^{i}(x)), vec(F_{S_{D}}^{i}(x))),
\end{equation}
where $x\in\mathbb{R}^{C\times H\times W}$ is an input image, $vec$ is the vectorization function, $F_{T}^{i}(x)=\frac{1}{4}\sum_{j=4i-1}^{4i+2}f_{T}^{j}(x)$, $F_{S_{D}}^{i}(x)=\frac{1}{4}\sum_{j=4i-3}^{4i} f_{S_{D}}^{j}(x)$, $T:\mathbb{R}^{C\times H\times W} \rightarrow \{f^{j}\in\mathbb{R}^{C'\times H'\times W'}\}_{j\in\{1, ..., 12\}}$, $S_{D}:\mathbb{R}^{C\times H\times W}\rightarrow \{f^{j}\in\mathbb{R}^{C'\times H'\times W'}\}_{j\in\{1, ..., 8\}}$, $f_{T}^{j}, f_{S_{D}}^{j}$ is the j-th level patch feature of the teacher and decoder student respectively. The mid-level layers of the teacher are found to be more robust to input noise and global representations, which are not suited to detecting local defects in industrial AD.

\subsection{Encoder-Encoder pair} 
The Encoder-Encoder model is designed to address semantic anomaly detection, where anomalies manifest as subtle, high-level irregularities within data. The techniques applied in industrial AD, such as alternative data flows through a decoder student model \cite{rd4ad, rd4ad++, uniad, mint-ad, dinomaly}, bottlenecks \cite{rd4ad, rd4ad++, dinomaly}, and dropout \cite{dinomaly}, effectively prevent the student model from learning an identity mapping but are less suitable for semantic AD. These methods convey information reduction, which impacts semantic anomaly detection. To mitigate this issue, we adopt an encoder-based student model, as in \cite{mkd, transformaly}. Inspired by the loose loss strategy proposed in \cite{dinomaly}, we introduce class token-wise distillation, replacing the conventional use of all patch tokens for distillation. Our encoder student model leverages class tokens from each layer of the teacher model, along with the final output class token. Since class token supervision is significantly sparser compared to full patch token supervision, it reduces the likelihood of the encoder student learning an identity mapping.

Both the teacher and student models are based on DINOv2’s encoder architecture. The teacher model, pre-trained on ImageNet, generates class tokens that encapsulate the high-level semantic content of input data. The student model, trained from scratch via distillation, learns to replicate these class tokens, embodying the semantic structure of data. Anomalies are detected by evaluating the discrepancy between the class tokens generated by the teacher and those by the student; a substantial discrepancy indicates that the input deviates from the normal patterns learned by the teacher, signaling semantic anomaly. Our loss function for the encoder student is following:
\vspace{-2mm}
\begin{equation}
L_{S_{E}}(x) = \frac{1}{m}\sum_{j=1}^{m} ||CLS_{T}^{j}(x)-CLS_{S_{E}}^{j}(x)||^{2},
\end{equation}
where $x$ is an input image, $T, S_{E}:\mathbb{R}^{C\times H\times W}\rightarrow \{CLS^{j}\in\mathbb{R}^{C'}\}_{j\in\{1,..., m\}}$, $CLS_{T}^{j}, CLS_{S_{E}}^{j}:\mathbb{R}^{C'}$ are the j-th level class token of the teacher and encoder student model respectively, $m$ is the number of ViT blocks, including the final class token. 

\subsection{Learning models}
To develop an unified general anomaly detection framework, we integrate the local anomaly score from the Encoder-Decoder model with the semantic anomaly score from the Encoder-Encoder model. These scores are combined using the Noisy-OR function \cite{milboost, mcboost} as \(P(x) = 1 - \frac{\exp(L_{S_{E}}(x))}{1+\exp(L_{S_{E}}(x))}\frac{\exp(L_{S_{D}}(x))}{1+\exp(L_{S_{D}}(x))}\), $P(x)$ denotes the probability that $x$ is normal. The probability gets close to 1, only if one of the two model scores is sufficiently small i.e. detecting normal samples.  The probability is low if both the models take large values, i.e. detecting anomalies. The total loss function is \(L_{total} = -\frac{1}{n}\sum_{i=1}^{n}\{y_{i}\log(P(x_{i}))+(1-y_{i})\log(1-P(x_{i}))\}\) 

where $x_{i}$ is the i-th input image, $y_{i}$ is the binary label for $x_{i}$ (1 for normal, 0 for abnormal) and $n$ is the number of training samples. The loss function takes the form of binary cross-entropy loss. In our experiments, only normal samples are accessible and used during the training phase, which aligns with the conventional AD settings. The proposed learning framework, however, can be extended to supervised settings where abnormal samples are also available during training.
Taking the derivative of \(L_{total}\), we observe the influence of co-training on each model. Let $L_{total}(x)$ be the loss of a normal sample $x$.
\begin{equation}
\frac{\partial L_{total}(x)}{\partial \theta_{S_{E}}} = \frac{\partial}{\partial \theta_{S_{E}}}\{-\log(P(x))\}= -\frac{1-P_{S_{D}}(x)}{P(x)}\frac{\partial P_{S_{E}}(x)}{\partial \theta_{S_{E}}},
\end{equation}
where $\theta_{S_{E}}$ denotes the parameters of the encoder student and $P_{S_{E}}(x)=\frac{1}{1+\exp(L_{S_{E}}(x))}$, $P_{S_{D}}(x)=\frac{1}{1+\exp(L_{S_{D}}(x))}$ is the probability that $x$ is normal predicted by the model $S_{E}$ and $S_{D}$, respectively. If the decoder student predict that \( x \) is normal with a high probability, the term \( \frac{1 - P_{S_{D}}(x)}{P(x)} \) becomes small, resulting in a reduced gradient for updating the encoder model and vice versa.

\cref{fig:tsne} illustrates the t-SNE \cite{tsne} visualization of feature distributions at different stages of learning, highlighting the discrepancy between the teacher and student models. Since our method identifies anomalies by measuring the difference between the teacher and student models, normal samples converge to the same point, whereas anomalous samples remain dispersed. In the initial phase (\cref{fig:tsne}(a), (c)), both the discrepancy between the teacher and the encoder student ($T$-$S_{E}$) and the discrepancy between the teacher and the decoder student ($T$-$S_{D}$) fail to effectively differentiate between normal and anomalous samples, exhibiting similarly scattered patterns for both. As training progresses (\cref{fig:tsne}(b), (d)), distinct separation patterns emerge. On the MVTec-AD dataset, $T$-$S_{E}$ struggles to cluster normal samples into a single point, whereas $T$-$S_{D}$ effectively separates normal and anomalous samples by consolidating normal samples into a distinct cluster. Conversely, on the CIFAR-10 dataset, the encoder student ($S_{E}$) progressively learns to distinguish the two classes, resulting in a well-defined separation between normal and anomalous instances. These observations underscore the complementary roles of the two student models in different anomaly detection scenarios.

\subsection{Inference}
To aggregate the anomaly detection results from the two KD models, 
the anomaly score is defined as \( AC(x) = 1- P(x) = \frac{\exp(L'_{S_{E}}(x))}{1+\exp(L'_{S_{E}}(x))}\frac{\exp(L_{S_{D}}(x))}{1+\exp(L_{S_{D}}(x))} \), where $x$ denotes the input image, \( S_{E} \) and \( S_{D} \) refer to the encoder student and decoder student models, respectively. 
Unlike the training phase, we use only the last layer class tokens for measuring the anomaly score of the encoder student \(L'_{S_{E}}(x) = ||CLS_{T}^{m}(x)-CLS_{S_{E}}^{m}(x)||^{2}\), where $CLS_{T}^{m}, CLS_{S_{E}}^{m}$ are the m-th (which is the last) class token of the teacher and encoder student respectively. Although the earlier layers of ViT aggregate features from other input patches, they retain less global structural information compared to the later layers. Note ViT is primarily trained using features from the final layer \cite{vit, dovit}. The use of only the last class token that captures high-level understanding has been shown more effective for semantic anomaly detection in experiments. By leveraging the strengths of both models through Noisy-OR, our ensemble approach achieves state-of-the-art performance across two distinct domains—industrial and semantic—under both single-class and multi-class settings, demonstrating robust generalization across diverse anomaly detection scenarios.

\section{Experiments}
\label{exp}
\subsection{Experiment Setup}
\label{subsec:experiment_setup}

\textbf{Datasets.} We conduct experiments on eight benchmark datasets: MVTec-AD \cite{mvtecad}, MVTec-LOCO \cite{mvtec-loco}, VisA \cite{visa}, Real-IAD \cite{real-iad} for industrial anomaly detection and CIFAR-10/100 \cite{cifar10}, Fashion-MNIST \cite{fashion-mnist}, View \cite{view} for semantic anomaly detection. 

\noindent \textbf{Evaluation Metrics.} We use the Area Under the Receiver Operating Characteristic Curve (AUROC) as the metric to evaluate anomaly detection performance. AUROC is widely used for binary classification tasks and is particularly useful for anomaly detection, as it measures the model ability to distinguish between normal and anomalous samples. Higher AUROC scores indicate better performances.

\noindent \textbf{Class settings.} To demonstrate the generalization of the proposed method we conduct experiments in two different AD settings; multi-class and single-class. In the \textbf{single-class setting}, we train a model for each class. We thus obtain as many models as classes and presume to know which class models to use, as in previous works. On the other hand, in the \textbf{multi-class setting}, a model is learned over all classes. The latter does not require much memory and prior-knowledge on classes, which is more general and easy to use in practice. For the multi-class semantic AD, experiments were done using four different splits per dataset, varying the normal class indices according to the settings of UniAD \cite{uniad}. In each split, half of the classes were designated as normal, while the remaining half were considered abnormal. The detailed class settings, including data splits for each dataset, are provided in \Cref{subsec:semanticAD_splits}.

\begin{table*}[ht]
\vspace{-3mm}
\caption{\textbf{Anomaly detection results under multi-class setting on on various datasets.} The best results in image-level AUROC(\%) are \textbf{bold} and the second best results are \underline{underlined}. If the original studies and the benchmark studies \cite{ader} did not provide results for certain datasets, these were marked as unavailable (’-’).}
\vspace{-3mm}
\label{tab:tab1}
\begin{center}
\resizebox{1\textwidth}{!}{
\begin{tabular}{cc|cccccccccc|c}
\hline
\multicolumn{2}{c|}{Category}                                                                     &KNN\cite{knn} &Transformaly\cite{transformaly}&MSC\cite{msc} & RD4AD\cite{rd4ad} & SimpleNet\cite{simplenet} & DeSTSeg\cite{destseg} & ReContrast\cite{recontrast} & UniAD\cite{uniad} & Dinomaly\cite{dinomaly} & GeneralAD\cite{generalad} & Ours   \\ \hline
\multicolumn{1}{c|}{\multirow{4}{*}{\rotatebox{90}{Industrial}}}                     & MVTec-AD   & 89.7     & 87.8                           &71.1            &94.6              & 95.3                      & 89.2                  & 98.3                        & 96.5              & \underline{99.6}                    &  56.8           & \textbf{99.7}   \\
\multicolumn{1}{c|}{}                                                                & VisA       & 81.1     & 65.0                        &-            &92.4              & 87.2                      & 88.9                  & 88.9                        & 88.8              & \underline{98.7}                    &  89.7            & \textbf{98.8}  \\
\multicolumn{1}{c|}{}                                                                & MVTec-LOCO & 74.8     & 57.2                        &-            &73.7                 & 81.8                      & 81.2                  &  -                          & 78.7               & 82.0                    &  \underline{82.4}           & \textbf{86.1}   \\
\multicolumn{1}{c|}{}                                                                & Real-IAD   & -        & -                              &-            & 82.7                 & 54.9                      & 79.3                  & 82.3                        & 83.1             & \textbf{89.3}           &  -                            & \underline{88.7}        \\ 
\hline
\multicolumn{1}{c|}{\multirow{4}{*}{\rotatebox{90}{Semantic}}}                       & CIFAR-10   & 92.1      & 93.6                           & 89.7         &-                 & -                         & -                     &  -                          & 87.2              & 71.5                    &  \underline{93.9}                         & \textbf{97.8}    \\
\multicolumn{1}{c|}{}                                                                & CIFAR-100  & 87.0     & 85.2                            & 83.7          &-                 & -                         & -                     &  -                          &  -               & 65.1                    &  \underline{89.7}                         & \textbf{92.3}    \\
\multicolumn{1}{c|}{}                                                                & FMNIST     & 88.2    & 85.6                             &  \underline{89.3}           &-                 & -                         & -                     &  -                          &  -                & 82.5                    &  \textbf{92.4}                & 88.3     \\
\multicolumn{1}{c|}{}                                                                & View       & 65.5     &\underline{82.2}               &66.6            &-                 & -                         & -                     &  -                          &  -                & 66.6                    &  70.5                         & \textbf{87.3}    \\ \hline
\end{tabular}%
}
\end{center}
\end{table*}

\begin{table*}[h]
\vspace{-3mm}
\caption{\textbf{Anomaly detection results under single-class settings in image-level AUROC(\%).} The best results in image-level AUROC(\%) are \textbf{bold} and the second best results are \underline{underlined}. 
Missing results from the original studies and the benchmark studies \cite{ader} for certain datasets are marked as (’-’).}
\vspace{-3mm}
\label{tab:tab2}
\begin{center}
\resizebox{1\textwidth}{!}{
\begin{tabular}{cc|cccccccccc|c}
\hline
\multicolumn{2}{c|}{Category}                                                  &KNN\cite{knn} &Transformaly\cite{transformaly}&MSC\cite{msc} & RD4AD\cite{rd4ad} & SimpleNet\cite{simplenet} & PatchCore\cite{patchcore} & ReContrast\cite{recontrast} & UniAD\cite{uniad} & Dinomaly\cite{dinomaly} & GeneralAD\cite{generalad} & Ours   \\ \hline
\multicolumn{1}{c|}{\multirow{4}{*}{\rotatebox{90}{Industrial}}} & MVTec-AD    & 90.2 & 84.0                          &87.2          & 98.4              & 99.6                      & 99.2                      & 99.5                        & 96.6              & \underline{99.7}           & 99.2                      & \textbf{99.8}   \\
\multicolumn{1}{c|}{}                                            & VisA        &-  & -                             & -             & 96.0              & 87.9                      & 94.2                      & 97.5                        & -                  & \underline{98.9}          & 95.9                      & \textbf{99.0}   \\
\multicolumn{1}{c|}{}                                            & MVTec-LOCO  &-  & -                             & -             & 79.7              & 77.6                      & 80.3                      & 82.1                        & -                  & 81.9                    & \underline{84.9}                      & \textbf{86.0}    \\
\multicolumn{1}{c|}{}                                            & Real-IAD    &-  & -                             & -            & -                 & -                         & 67.2                      & -                           & -                  & \textbf{92.0}                    & -                                       & \underline{91.3}       \\ 
\hline
\multicolumn{1}{c|}{\multirow{4}{*}{\rotatebox{90}{Semantic}}}   & CIFAR10     &97.7 & 98.3                          &97.5          & 86.5              & 86.5                      & 64.1                      & 84.1                        & -                  & 90.4                    & \textbf{99.3}                      & \textbf{99.3}        \\
\multicolumn{1}{c|}{}                                            & CIFAR-100   &97.0 & \underline{97.3}              &96.4          & -                 & -                         & -                         & 84.0                        &  -                & 90.3                    & \textbf{98.4}              & 97.2     \\
\multicolumn{1}{c|}{}                                            & FMNIST      &94.2 & 94.4                          & \underline{95.0}            & \underline{95.0}              & 87.4                      & 77.4                      & 92.4                        & -                  & -                       & \textbf{95.2}             & 94.7       \\
\multicolumn{1}{c|}{}                                            & View        &94.6 & 95.8                          & 95.1             & -                 & 76.8                      & -                         & -                           & -                  & -                       & \underline{95.9}                      & \textbf{96.5}     \\ \hline
\end{tabular}%
}
\end{center}
\vspace{-5mm}
\end{table*}

\subsection{Results}
\textbf{Industrial Defect Detection.} We evaluate our model's performance on industrial anomaly detection using the MVTec-AD, MVTec-LOCO, VisA, and Real-IAD datasets, benchmarking over state-of-the-art methods, including RD4AD \cite{rd4ad}, SimpleNet \cite{simplenet}, DeSTSeg \cite{destseg}, ReContrast \cite{recontrast}, UniAD \cite{uniad}, Dinomaly \cite{dinomaly}, and GeneralAD \cite{generalad}. For the methods where multi-class AD accuracies are not reported in their original works, we referenced the results from the benchmark studies \cite{ader,dinomaly} to ensure  comprehensive comparisons.

As shown in the upper section of \cref{tab:tab1} and \cref{tab:tab2}, the proposed method achieves consistently high AUROC scores across all four industrial datasets at both single-class and multi-class settings. Notably, we obtain a significant accuracy gain compared to Dinomaly on MVTec-LOCO. The dataset includes both minor defects and logical anomalies which is hard to be dealt with by prior specialist methods. This improvement supports that the two student models in our framework jointly tackles complex and mixed anomaly types. Detailed class-wise evaluation results for MVTec-AD are presented in \cref{tab:MVTec-AD}.

\noindent\textbf{Semantic Anomaly Detection.} We evaluate our model on the CIFAR-10/100, Fashion-MNIST, and View datasets for semantic anomaly detection, in comparison to state-of-the-art methods such as KNN \cite{knn}, Transformaly \cite{transformaly}, MSC \cite{msc}, UniAD \cite{uniad}, Dinomaly \cite{dinomaly}, and GeneralAD \cite{generalad}. As shown in the lower section of \cref{tab:tab1} and \cref{tab:tab2}, our approach achieves high AUROC scores across all four semantic anomaly detection datasets. Detailed evaluation results on CIFAR-10 with varying normal indices are presented in \cref{tab:cifar10}.
While the CIFAR-10/100 and View datasets capture natural images, FMNIST—a small grayscale image dataset (28x28)—differs significantly from natural images, making it difficult to maintain consistently high performance across diverse semantic anomaly detection datasets. Interestingly, as shown in \cref{tab:tab1} and \cref{tab:tab2}, the methods specializing in industrial anomaly detection generally perform better on FMNIST than on other semantic AD datasets. Note our method demonstrates 
lower performance variance as well as the highest mean performance across the four semantic AD datasets. The decoder student in our method, which focuses on industrial AD, enhances performance on FMNIST. By leveraging the strengths of both models, our dual approach effectively addresses these complex anomaly types.

\noindent\textbf{Discussions.}
The experimental results demonstrate that the proposed method outperforms existing SOTA methods in both industrial and semantic anomaly detection. The key to this success lies in the dual-model ensemble, where the Encoder-Encoder model specializes in detecting high-level semantic anomalies, and the Encoder-Decoder model excels at capturing fine-grained structural differences. We also observe that the accuracy of GeneralAD on MVTec-AD drops drastically when applied to multi-class settings. The results show that in GeneralAD the choice of anomaly synthesis strategy is crucial for their performance. On the other hand, our method shows robust performance without any dataset-specific knowledge and different anomaly types at both multi-class and single-class settings. 

\subsection{Ablation study}
\label{sec:ablation}
To better understand the contributions of each component within our ensemble framework, an ablation study is performed by evaluating the performance of partial models and comparing these results with the comprehensive version of our method. \cref{tab:ablation} presents the results, where we evaluate multi-class anomaly detection performance on MVTec-AD, MVTec-LOCO, and CIFAR-10 by varying three experimental conditions. The first two conditions, \( S_{E} \) and \( S_{D} \), assess the effectiveness of each of the student models in our dual-model framework. The third option, \( CLS^{m} \), indicates whether the final class token is used for inference. When \( CLS^{m} \) is not selected, we calculate \( AC(x) \) using  \( L''_{S_{E}}(x) \) instead of \( L'_{S_{E}}(x) \) where \(L''_{S_{E}}(x)=\frac{1}{m-1}\sum_{i=1}^{m-1}||CLS_{T}^{j}(x)-CLS_{S_{E}}^{j}(x)||^{2}\). 
\begin{wraptable}{r}{0.5\textwidth}
  \centering
    \caption{Effect of each component in the proposed method on MVTec-AD, MVTec-LOCO and CIFAR-10. $L_{S_{E}}$: Loss function of encoder student. $L_{S_{D}}$: Loss function of decoder student. $CLS^{m}$: Using the last class token for inference. Noisy-OR: Using Noisy-OR ensemble approach. The best results are \textbf{bold} and the second best results are \underline{underlined}.}
  \centering
   \resizebox{0.5\textwidth}{!}{
    \begin{tabular}{cccc|ccc}
\hline
\multirow{2}{*}{$L_{S_{E}}$}     & \multirow{2}{*}{$L_{S_{D}}$}       & \multirow{2}{*}{$CLS^{m}$} & \multirow{2}{*}{Noisy-OR}         & \multirow{2}{*}{MVTec-AD}      & \multirow{2}{*}{MVTec-LOCO}    & \multirow{2}{*}{CIFAR-10}      \\
                             &                                &                            &                                     &                                        \\ \hline 
\checkmark                   &                                &                            & &   96.46                             & 78.09                                      & 85.84                          \\
                             & \checkmark                     &                            & &   99.64                             & 82.51                                       & 71.02                        \\
\checkmark                   &                                & \checkmark                 & &   96.58                             & 79.23                                     & 97.71                         \\
\checkmark                   & \checkmark                     &                            & \checkmark &  99.65                  & 84.86                          &  83.92                        \\
\checkmark                   & \checkmark                     & \checkmark                 &  &  \textbf{99.73}                     & \underline{85.65}                             & \textbf{97.90}                 \\
\checkmark                   & \checkmark                     & \checkmark                 & \checkmark &  \underline{99.66}                     & \textbf{86.10}                             & \underline{97.80}                 \\ \hline
\end{tabular}
}
\label{tab:ablation}
    \label{tab:few_shot}
    \vspace{-4mm}
\end{wraptable}
The last option, Noisy-OR stands for enabling Noisy-OR ensemble approach. When Noisy-OR is disabled, we simply sum \( L'_{S_{E}}(x) \) and \( L_{S_{D}}(x) \) to aggregate outputs.

The ablation study results emphasize the significance of combining both the Encoder-Encoder and Encoder-Decoder models. While the Encoder-Encoder model performs effectively on CIFAR-10, its performance on MVTec-AD is lower than that of the Encoder-Decoder model. Conversely, the Encoder-Decoder model excels at industrial anomaly detection but underperforms in semantic tasks. When combined, the ensemble achieves state-of-the-art performance across both datasets. The results also demonstrate that utilizing the last class token for inference notably enhances the encoder student model, particularly on the CIFAR-10 semantic AD dataset. Note that Noisy-OR is particularly effective for MVTec-LOCO which have complex types of anomalies (e.g., logical anomalies).
We also report the enhanced performance of previous methods in \cref{tab:ablation_DINOv2} to ensure a fair comparison. Our method achieves the best performance among all AD methods when the same backbone network is applied. Furthermore, in \cref{tab:ablation_backbone}, we evaluate the complexity of our method using two different backbones and compare it with GeneralAD to demonstrate the efficiency of our proposed approach. In these experiments, we use the single-class setting for GeneralAD. Our method achieves comparable results to GeneralAD, despite GeneralAD utilizing class knowledge in the single-class setting. Since the single-class setting requires class-wise training, the number of parameters in GeneralAD increases by a factor of $n$, where $n$ is the number of classes in the dataset. Additionally, our method demonstrates reasonable AD performance even when using a smaller backbone network with fewer parameters and faster inference speed.

\vspace{-3mm}
\begin{table}[ht]
  \centering
  \begin{minipage}{0.48\linewidth}
   \centering
    \caption{\textbf{Comparison of our method with other state-of-the-art methods using DINOv2 backbone.}}
    \label{tab:ablation_DINOv2}

   \resizebox{1\columnwidth}{!}{
    \begin{tabular}{c|c|ccc}
\hline
\multirow{2}{*}{Method}       & \multirow{2}{*}{Backbone}       &  \multirow{2}{*}{MVTec-AD}      & \multirow{2}{*}{MVTec-LOCO}    & \multirow{2}{*}{CIFAR-10}      \\
                             &                                  &                                &                                 &                                 \\     \hline 
RD4AD                       &  WideResNet50                     &   94.6                         & 73.7                            &  47.7                         \\
                            &  DINOv2-B-reg4/14                 &   98.4                         & 81.6                            &  74.7                          \\  \hline 
Transformaly                &  ViT-B/16                         &   87.8                         & 57.2                            &  93.6                     \\
                            &  DINOv2-B-reg4/14                 &   90.0                         & 82.6                            &  \textbf{98.2}                       \\     \hline 
GeneralAD                   &  DINOv2-B-reg4/14                 &   56.8                         & 82.4                            &  93.9                      \\      \hline 
Dinomaly                    &  DINOv2-B-reg4/14                 &   99.6                         & 82.0                            &  71.5                        \\     \hline    
Ours                        &  DINOv2-B-reg4/14                 &   \textbf{99.7}                         & \textbf{86.1}                            &  97.8      \\ \hline 
\end{tabular}
}
  \end{minipage}
 \hfill
  \begin{minipage}{0.48\linewidth}
  \vspace{-3mm}
  \tabcolsep=0.08cm
\caption{\textbf{Complexity comparison between our method and GeneralAD.} $n$ is the number of classes, 15 for MVTec-AD, 5 for MVTec-LOCO and 5 for CIFAR-10.}

\label{tab:ablation_backbone}
  \centering
   \resizebox{1\columnwidth}{!}{
    \begin{tabular}{c|c|c|c|ccc}
\hline
\multirow{2}{*}{Method}     & \multirow{2}{*}{Backbone}   & \multirow{2}{*}{\shortstack{Param\\(M)}}    & \multirow{2}{*}{\shortstack{Inference\\Time(ms)}}   &  \multirow{2}{*}{\shortstack{MVTec\\-AD}}      & \multirow{2}{*}{\shortstack{MVTec\\-LOCO}}    & \multirow{2}{*}{CIFAR-10}       \\
                            &                            &                            &                                   \\ \hline 
GeneralAD                   & DINOv2-B-reg4/14           &    $93.1\times n$                  &   24.932                  &99.2 &84.9 &99.3        \\   \hline 
Ours                        & DINOv2-S-reg4/14           &    58.7                  &   65.592                 &   98.8                 & 81.7                            &  93.3                                           \\
                            & DINOv2-B-reg4/14           &    233.1                 &   93.164       &   99.7                         & 86.1                            &  97.8                            \\ \hline

\end{tabular}
}
\end{minipage}
\end{table}
\vspace{-3mm}

\section{Few-shot AD scenarios}

\begin{wraptable}{r}{0.4\textwidth}
  \centering
  \vspace{-5mm}
\caption{\textbf{Few-shot AD performance on MVTec-AD.} Shots indicates the number of normal data samples used in training.}
    \resizebox{0.4\textwidth}{!}{
    \begin{tabular}{lcccc}
        \toprule
         Shots & 1 & 2 & 4 & 8 \\
        \midrule
        SPADE \cite{spade} & 71.6 & 73.4 & 82.8 & 84.0\\
        PaDiM \cite{padim} & 76.1 & 78.9 & 80.5 & 82.0\\
        PatchCore \cite{patchcore} & 84.1 & 87.2 & 88.5 & 92.2\\
        GeneralAD \cite{generalad}& \underline{87.5} & \underline{91.5}& \underline{92.8}& \underline{93.6}\\
        \midrule
        Ours & \textbf{92.6} & \textbf{94.6}& \textbf{94.9}& \textbf{96.3}\\
        \bottomrule
    \end{tabular}}
    \label{tab:few_shot}
    \vspace{-5mm}
\end{wraptable}

In \cref{tab:few_shot}, our approach demonstrates strong performance in few-shot anomaly detection scenarios, where abnormal data is unavailable, and even normal data is scarce. This setting is closely aligned with practical, real-world applications. For the methods that do not report few-shot AD performance in their original works, we refer to the few-shot experiments of \cite{generalad} for fair comparisons. We evaluate the single-class anomaly detection performance on MVTec-AD by varying the number of shots for training the models. Our method surpasses these baseline approaches by a substantial margin, demonstrating that our model achieves high sample efficiency with limited training samples. As in \cite{adaclip, anomalyclip, inctrl}, leveraging the strong prior of of vision-language Models (VLMs) has the potential to further enhance the performance. 

\section{Conclusions}
\label{conclusions}

In this paper, we propose a novel approach to general anomaly detection that integrates two knowledge distillation (KD)-based teacher–student models: an Encoder–Decoder model for industrial anomaly detection and an Encoder–Encoder model for semantic anomaly detection. By incorporating dedicated branches for industrial and semantic anomalies, the method supports effective training on heterogeneous datasets and enables domain-agnostic inference without prior knowledge of the test domain. The effectiveness of our dual-model architecture and the contribution of each component are validated through extensive experiments on eight public benchmarks and ablation studies. The proposed ensemble method achieves state-of-the-art performance on both industrial and semantic anomaly detection tasks, while also demonstrating strong generalization in multi-class settings involving complex datasets with mixed anomaly types. 

\textbf{Limitations.} The limitations of our method are discussed in \Cref{sec:limitaions}.

\section*{Acknowledgment}
This work was supported by NST grant (CRC 21011, MSIT), IITP grant (RS-2023-00228996, RS-2024-00459749, RS-2025-25443318, RS-2025-25441313, MSIT) and KOCCA grant (RS-2024-00442308, MCST).

{
\small
\bibliographystyle{unsrt}
\bibliography{main}
}

\clearpage
\appendix

\section{Broader Impacts}
\label{sec:broader_impacts}
Our model bridges semantic anomalies and industrial defect detection tasks in a unified architecture, reducing the need for task-specific models and training. This enables efficient deployment in real-world scenarios, such as smart factories and safety-critical applications, while lowering operational and computational costs. Furthermore, the generality of our approach facilitates wider adoption in low-resource environments, contributing to the wider dissemination and practical availability of anomaly detection technologies.

\section{Limitations}
\label{sec:limitaions}
\begin{figure}[ht]
\vspace{-3mm}
\begin{center}
\includegraphics[width=0.25\columnwidth]{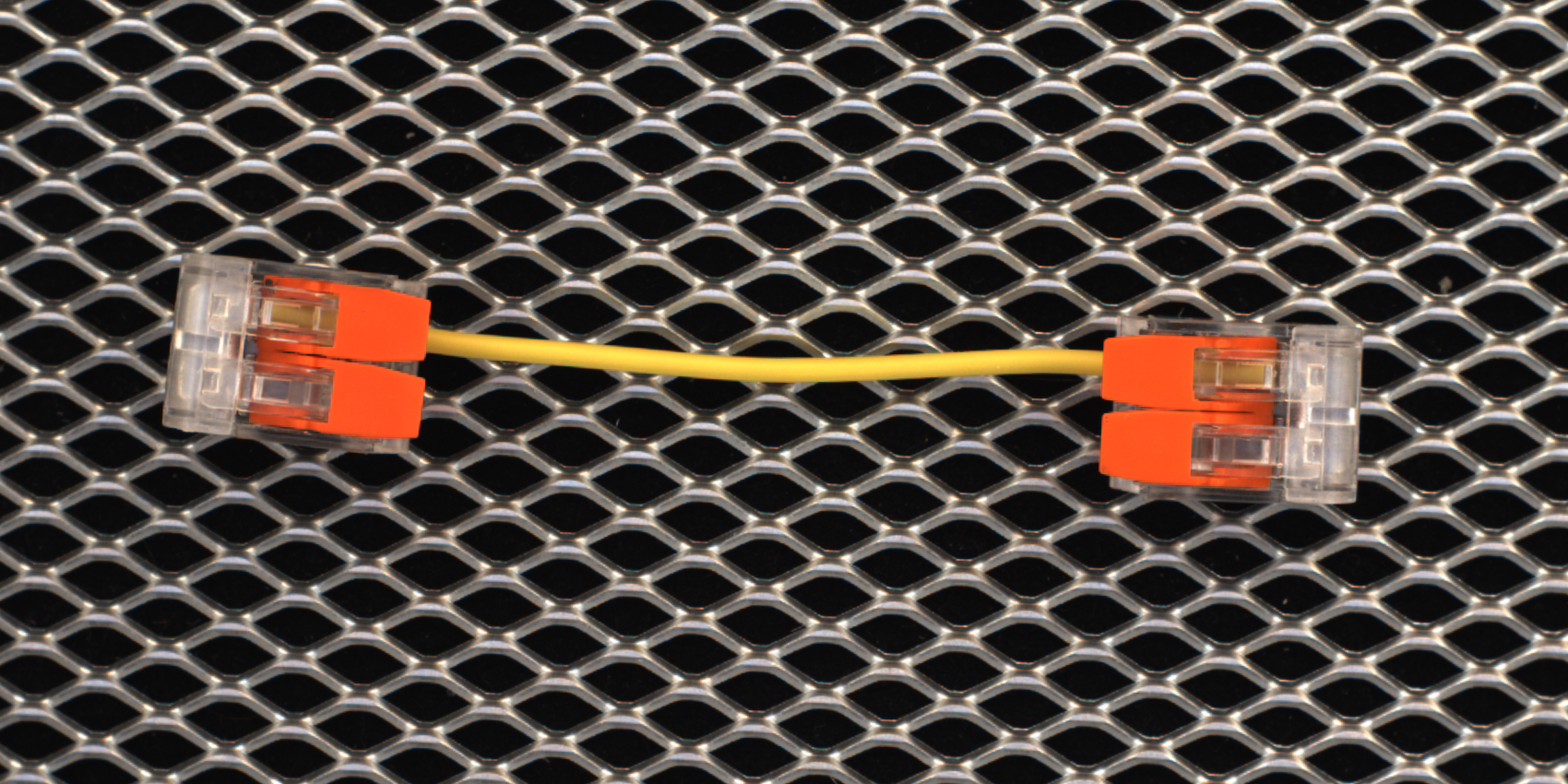}
\includegraphics[width=0.25\columnwidth]{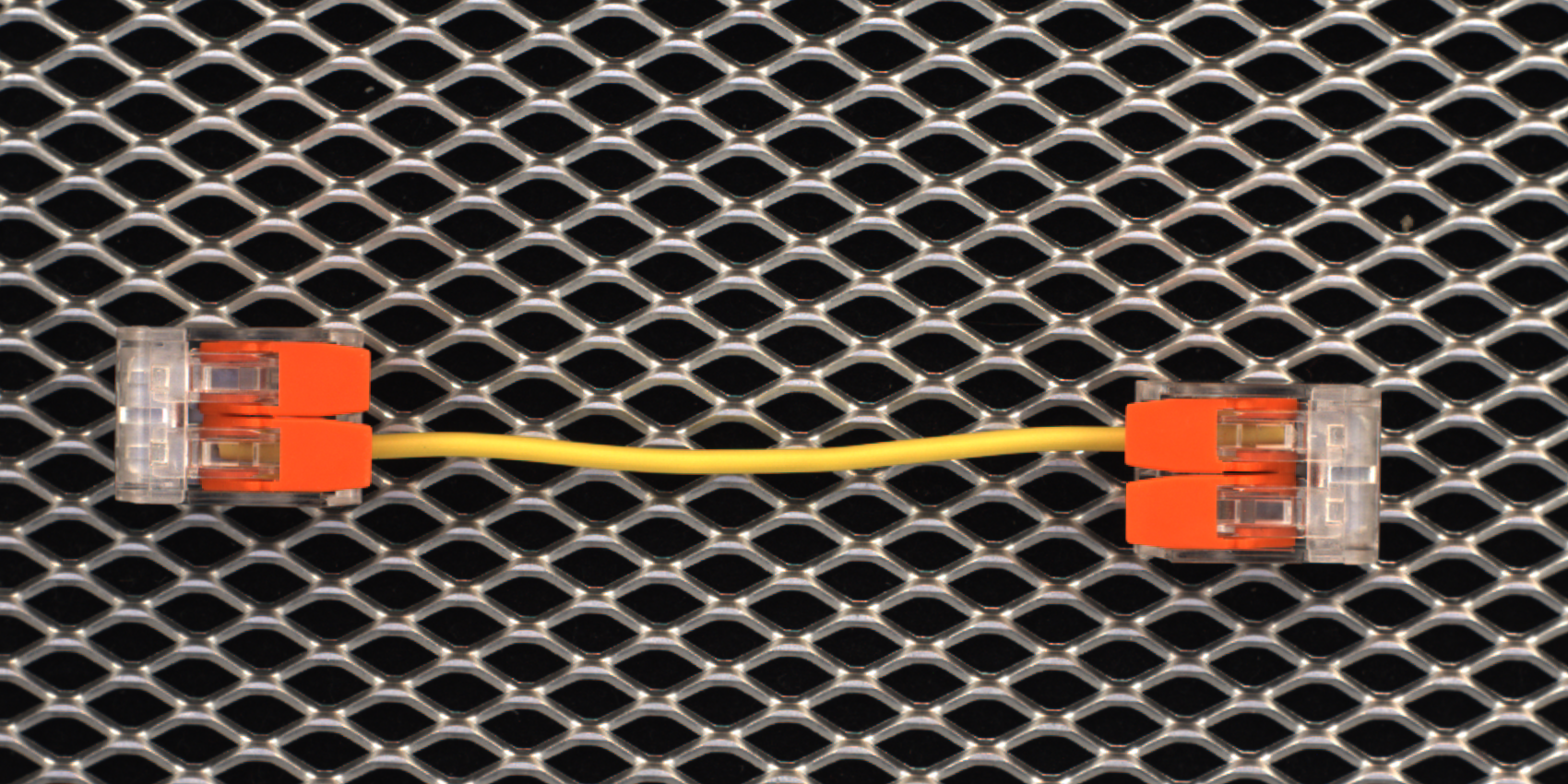}
\includegraphics[width=0.25\columnwidth]{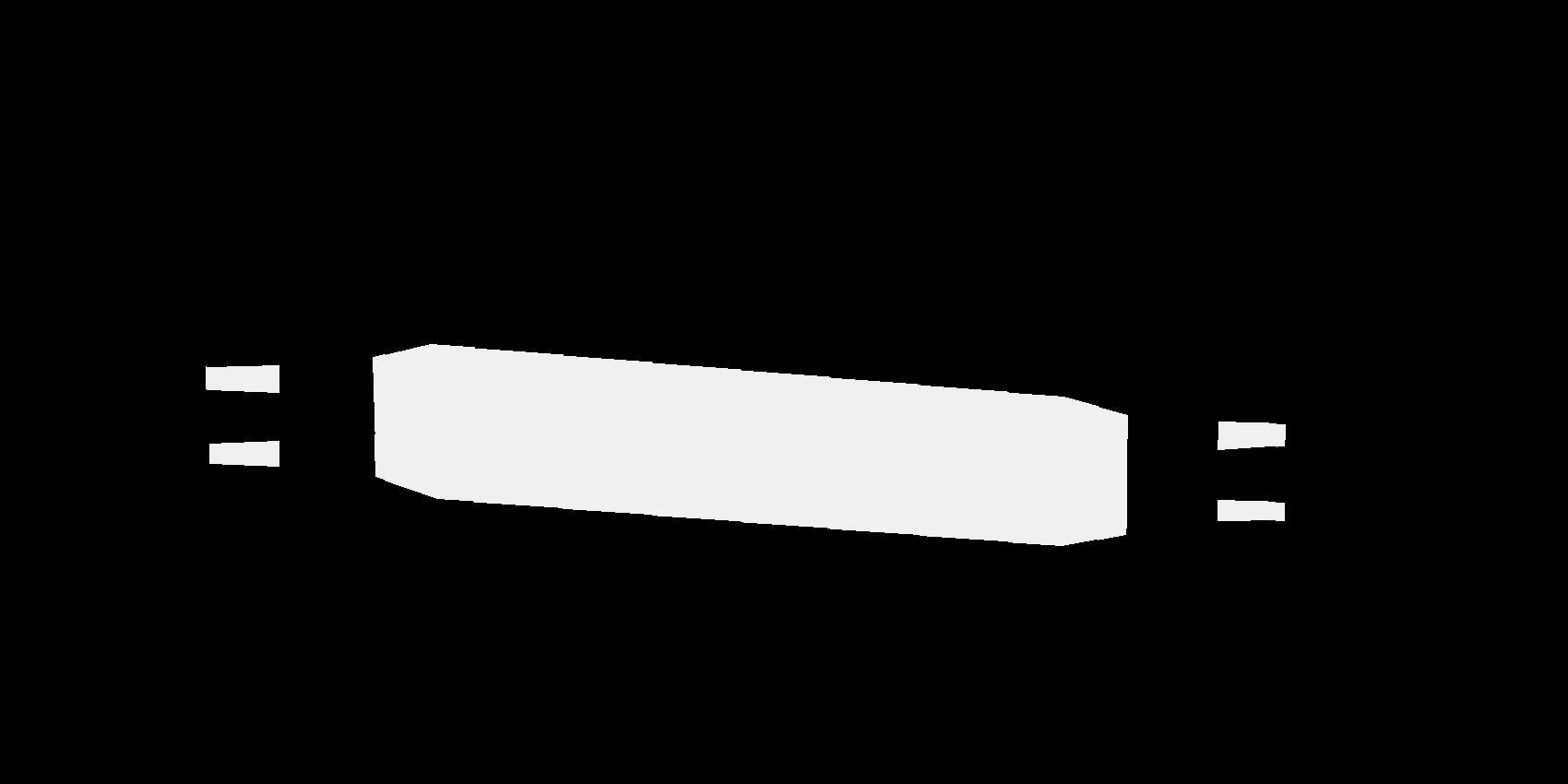}
\includegraphics[angle=90, width=0.25\columnwidth]{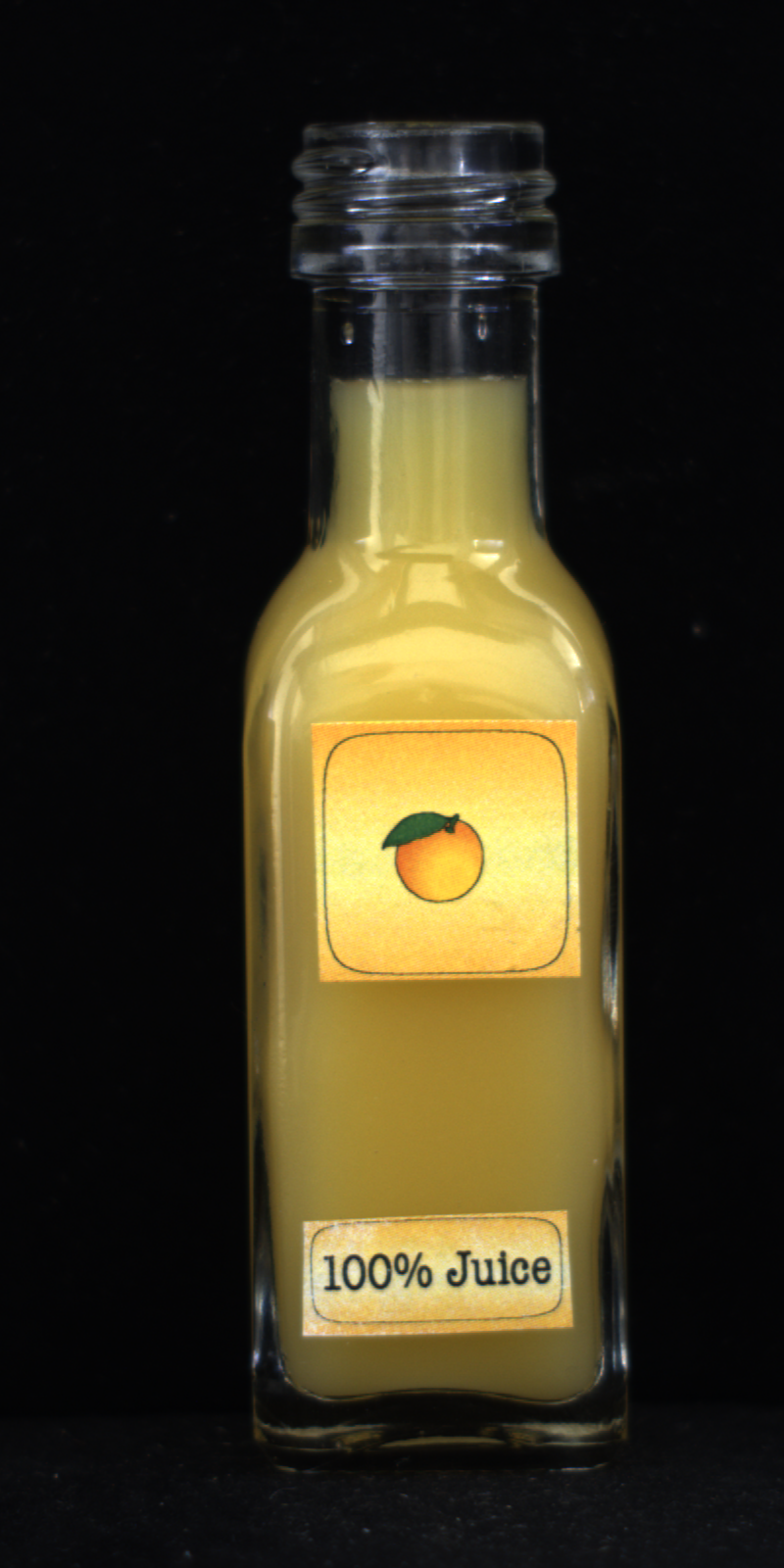}
\includegraphics[angle=90, width=0.25\columnwidth]{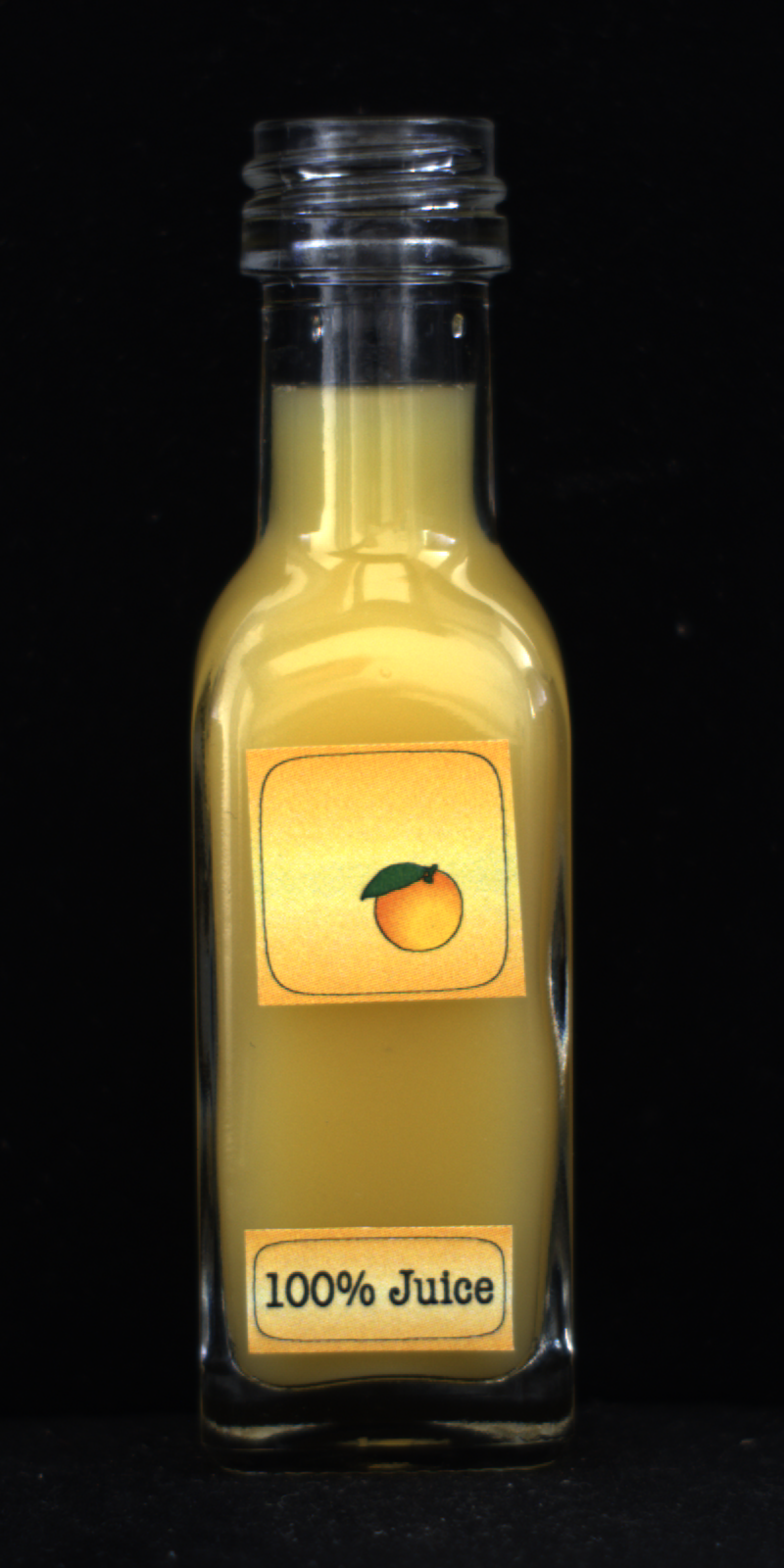}
\includegraphics[angle=90, width=0.25\columnwidth]{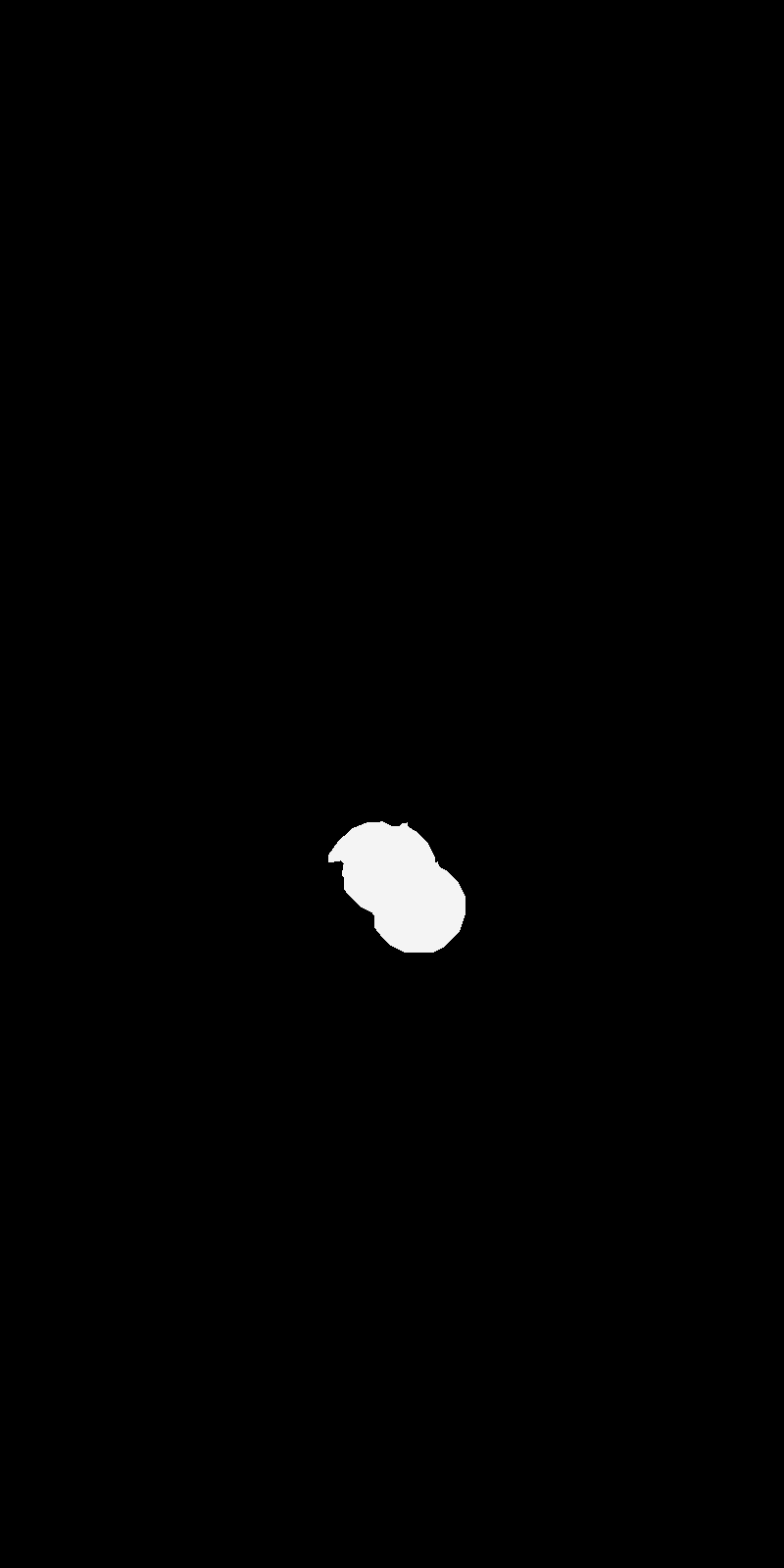}
\includegraphics[width=0.25\columnwidth]{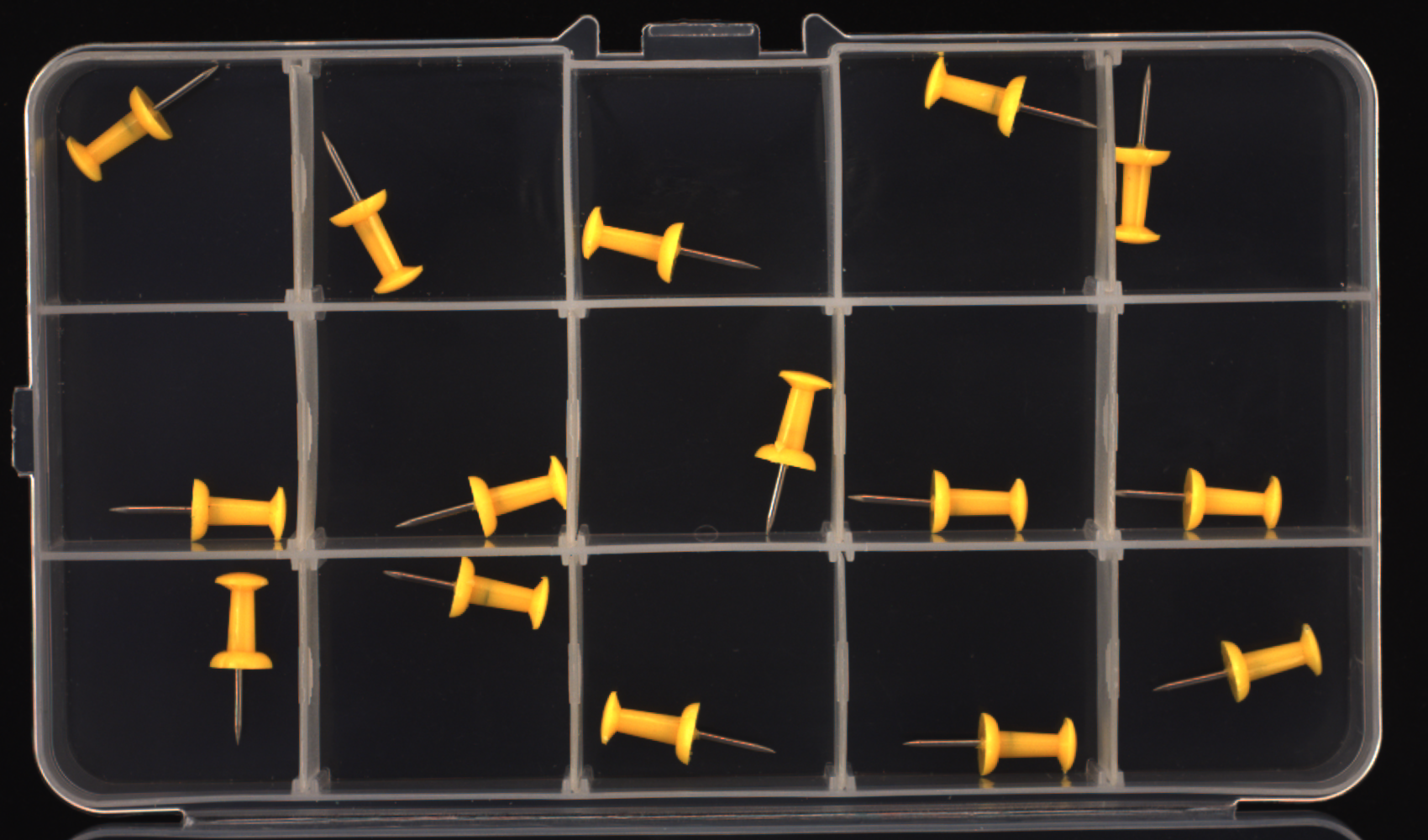}
\includegraphics[width=0.25\columnwidth]{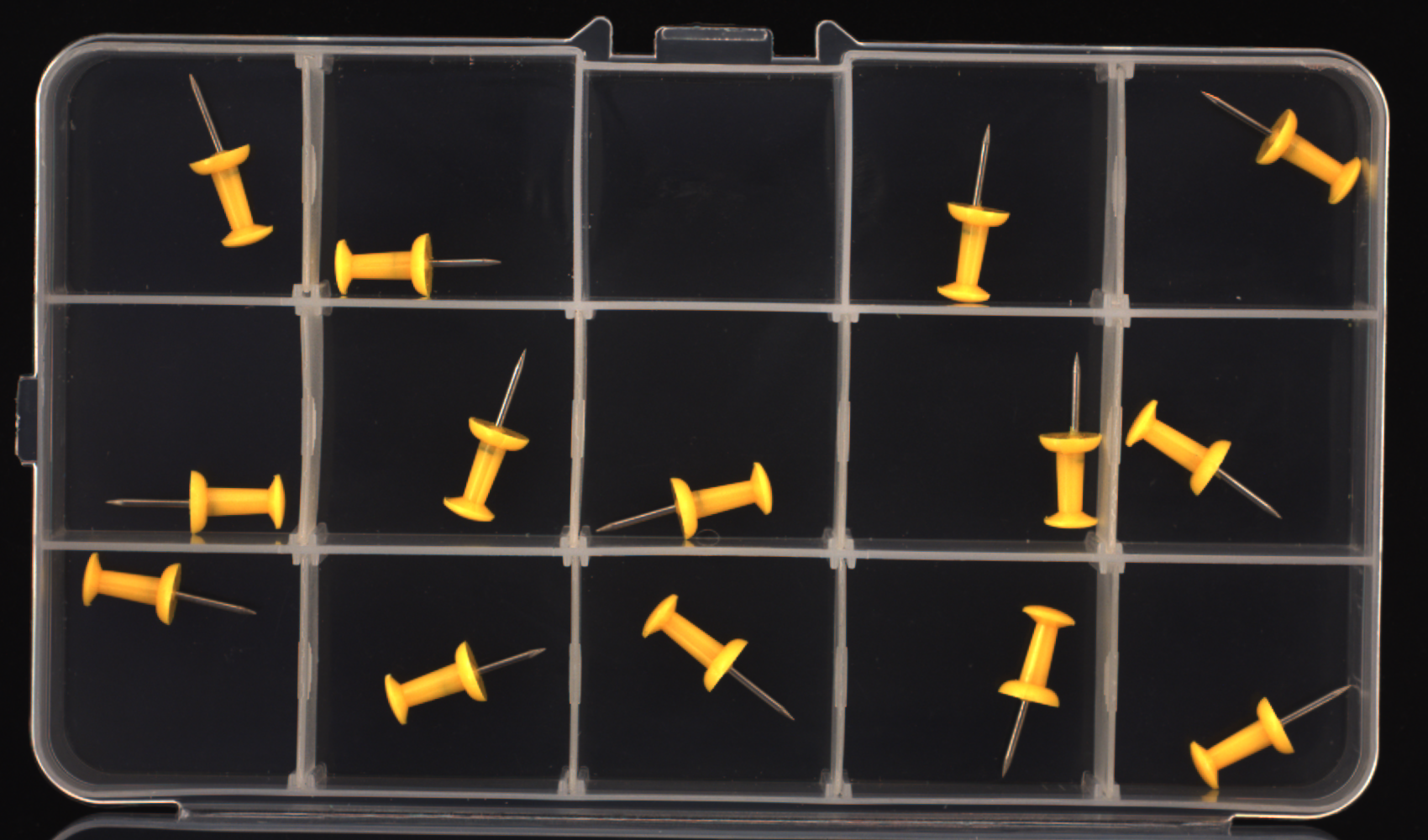}
\includegraphics[width=0.25\columnwidth]{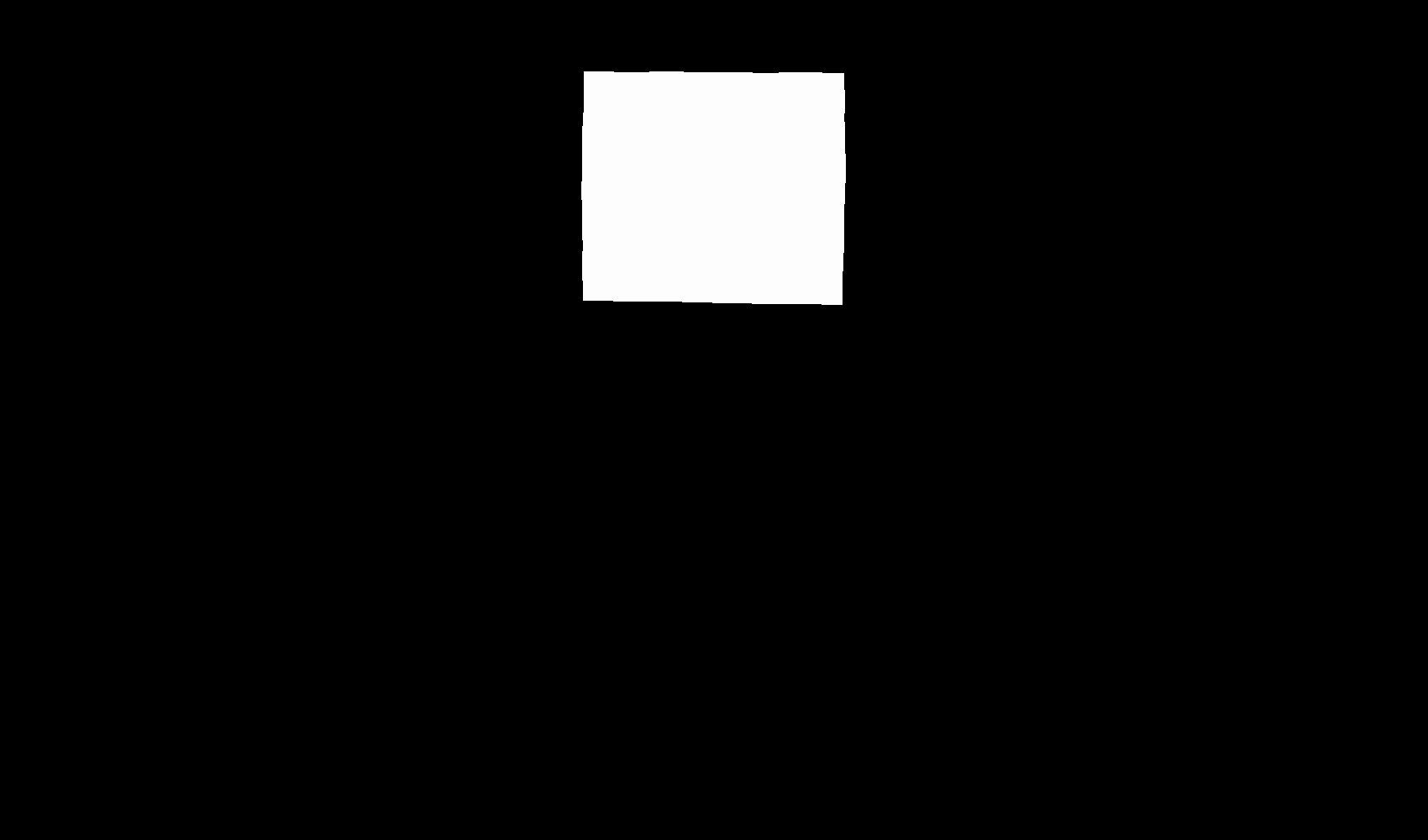}
\end{center}
\vspace{-3mm}
\caption{\textbf{Failure cases in MVTec-LOCO dataset.} (Left) normal (Middle) logical anomaly sample, (Right) ground truth anomaly mask}
\label{fig:limitation}
\vspace{-3mm}
\end{figure}

The proposed method employs two explicit student models trained simultaneously, which increases model complexity and inference time. We believe that a routing network \cite{moead} could effectively mitigate these issues. A failure example from MVTec-LOCO is shown in \cref{fig:limitation}. This logical and local anomaly is difficult to detect using patch-based analysis, since it does not show explicit defect patches. Note this sample also does not significantly alter the semantic content of the image. Consequently, it challenges our model's reliance on patch-level and semantic-level analysis. Further improvements are needed to address such types of anomalies.

\section{Datasets}

\textbf{MVTec-AD} \cite{mvtecad} MVTec-AD consists of 5354 high-resolution images from 15 different product categories, such as bottles, screws, and transistors, with annotated anomalies like scratches, dents, and missing parts. 

\noindent\textbf{MVTec-LOCO} \cite{mvtec-loco} MVTec-LOCO contains images of industrial products that exhibit logical and structural defects. It comprises 3,644 images across five classes.

\noindent\textbf{VisA} \cite{visa} Visual Anomaly (VisA) dataset is a benchmark for industrial anomaly detection, with images capturing diverse manufacturing defects across 12 classes in three distinct domains, totaling 10,821 high-resolution images. 

\noindent\textbf{Real-IAD} \cite{real-iad} Real-IAD is a large, recently published dataset for unsupervised anomaly detection (UAD), featuring 30 different objects. Using the official data split, the dataset includes 36,465 normal images in the training set and 114,585 images in the test set, of which 63,256 are normal and 51,329 are anomalous. 

\noindent\textbf{CIFAR-10/100} \cite{cifar10} CIFAR-10 and CIFAR-100 datasets consist of 60,000 natural color images at a resolution of 32$\times$32 pixels, with 50,000 images designated for training and 10,000 for testing. CIFAR-10 includes 10 classes of equal size, while CIFAR-100 is organized into 100 fine-grained classes or 20 broader classes. In our experiments, we use the 20 coarse-grained classes. 

\noindent\textbf{Fashion-MNIST} \cite{fashion-mnist} Fashion-MNIST provides a total of 60,000 training samples and 10,000 test samples, each represented as a 28$\times$28 grayscale image, divided into 10 distinct classes. 

\noindent\textbf{View} \cite{view} The View dataset consists of natural scene images organized into six categories; buildings, forest, glacier, mountain, sea and street. It includes roughly 2,300 images per category for training and about 500 images per category for testing.

\subsection{Details on Split for Multi-Class Semantic Anomaly Detection}
\label{subsec:semanticAD_splits}
For multi-class semantic anomaly detection, we construct four splits per dataset, with each split comprising half of the classes selected as normal. The indices of the normal classes for each split are provided in \cref{tab:multi_class_cifar100_indices}, \cref{tab:multi_class_fmnist_indices}, \cref{tab:multi_class_view_indices}.

\begin{table}[h]
\vspace{-3mm}
\begin{minipage}{0.48\linewidth}
\centering
\caption{\textbf{Details on split for multi-class anomaly detection on Fashion-MNIST.}}
{
\begin{tabular}{@{}c|c@{}}
\toprule
\textbf{Split} & \textbf{Normal indices}                                                                                                         \\ \midrule
A              & \{0, 1, 2, 3, 4\}  \\ \midrule
B              & \{5, 6, 7, 8, 9\}   \\ \midrule
C              & \{0, 2, 4, 6, 8\} \\ \midrule
D              & \{1, 3, 5, 7, 9\}   \\  \bottomrule
\end{tabular}
}
\label{tab:multi_class_fmnist_indices}
\end{minipage}
\hfill
\begin{minipage}{0.48\linewidth}
\centering
\caption{\textbf{Details on split for multi-class anomaly detection on View.}}
{
\begin{tabular}{@{}c|c@{}}
\toprule
\textbf{Split} & \textbf{Normal classes}                                                                                                         \\ \midrule
A              & \{buildings, forest, glacier\}  \\ \midrule
B              & \{mountain, sea, street\}   \\ \midrule
C              & \{buildings, glacier, sea\} \\ \midrule
D              & \{forest, mountain, street\}   \\  \bottomrule
\end{tabular}
}
\label{tab:multi_class_view_indices}
\end{minipage}
\vspace{-3mm}
\end{table}

\begin{table}[h]
\centering
\caption{\textbf{Details on split for multi-class anomaly detection on CIFAR-100.}}
{
\begin{tabular}{@{}c|c@{}}
\toprule
\textbf{Split} & \textbf{Normal indices}                                                                                                         \\ \midrule
A              & \{0, 1, 2, 3, 4, 5, 6, 7, 8, 9\}  \\ \midrule
B              & \{10, 11, 12, 13, 14, 15, 16, 17, 18, 19\}   \\ \midrule
C              & \{0, 2, 4, 6, 8, 10, 12, 14, 16, 18\} \\ \midrule
D              & \{1, 3, 5, 7, 9, 11, 13, 15, 17, 19\}   \\  \bottomrule
\end{tabular}
}
\label{tab:multi_class_cifar100_indices}
\vspace{-3mm}
\end{table}
\section{Implementation Details}
\label{sec:implementation_details}

We utilize the ViT-Base/14 architecture as the teacher encoder, student encoder, and student decoder. The teacher encoder is pre-trained using DINOv2-Reg4 \cite{dinoreg}. The default drop rate of the Noisy Bottleneck is set to 0.2 and  increased to 0.4 when applied to the Real-IAD dataset.
For the industrial dataset, input images are resized to 448 $\times$ 448 and then center-cropped to 392 $\times$ 392, while for the semantic dataset, images are resized and center-cropped to 392 $\times$ 392.
The StableAdamW optimizer \cite{stableadam} with AMSGrad \cite{amsgrad} is employed, with a learning rate of 2e-3, $\beta$=(0.9, 0.999), a weight decay of 1e-4, and $\epsilon$=1e-10 for the industrial dataset, while a learning rate of 1e-4 is used for the semantic dataset.
The networks are trained for 50,000 iterations on MVTec-AD, MVTec-LOCO, VisA and Real-IAD, and 82,500 iterations on CIFAR-10, and 62,500 iterations on CIFAR-100, FMNIST and View in a multi-class setting. For single-class experiments, the networks are trained for 5,000 iterations on all datasets. The batch size is fixed at 8 across all configurations. 
The implementation is based on Python 3.8 and PyTorch 1.12.0 with CUDA 11.3. Experiments were run on an NVIDIA GeForce RTX 3090 GPU (24GB).

\section{Ablation Study for GeneralAD}
\label{sec:sup_ablation_generalAD}
We evaluate single-class anomaly detection performance in terms of image-level AUROC(\%) by varying the anomalous feature generation strategies of GeneralAD.

\begin{table}[h]
\caption{\textbf{Effect of Anomalous Feature Generation Strategies of GeneralAD on MVTec-AD, MVTec-LOCO and CIFAR-10.} The best results are \textbf{bold}.}
  \centering
   \resizebox{1\columnwidth}{!}{
    \begin{tabular}{l|ccc}
\hline
\textbf{Strategy}& MVTec-AD      & MVTec-LOCO    & CIFAR-10      \\ \hline 
Noise Random Patches&   \textbf{99.2}                             & 83.9                                      & 96.3                          \\
Noise Random Patches and Attention Shuffle&   98.9                             & \textbf{84.9}                                       & 93.2                        \\
Noise All Patches&   99.1                             & 83.7                                     & \textbf{99.3}                         \\ \hline
\end{tabular}
}
\label{tab:generalad_ablation}
\vspace{-5mm}
\end{table}

\section{Comparison of Class Token and Patch Token Distillation for Encoder-Encoder Model}
In \Cref{tab:token_ablation}, we present an ablation study comparing class token and patch token distillation for the Encoder-Encoder model. Employing the class token for knowledge distillation yields a 3.80\% performance improvement on CIFAR-10 over the patch token distillation.

\begin{table}[h]
\vspace{-3mm}
\caption{\textbf{Comparison of class token and patch token distillation for our Encoder-Encoder model.} Anomaly detection results in AUROC on CIFAR-10 under multi-class setting. The best results are \textbf{bold}.}
  \centering
   {
    \begin{tabular}{l|c}
\hline
\multirow{2}{*}{Token for Distillation}        & \multirow{2}{*}{CIFAR-10}      \\
                             &                                    \\ \hline 
Patch Token                                   & 94.00                                \\
Class Token (Ours)                               & \textbf{97.80}                 \\ \hline
\end{tabular}
}
\label{tab:token_ablation}
\vspace{-4mm}
\end{table}

\section{Comparison with Prior Works} 
The most relevant works include Dinomaly (the SOTA multi-class AD method on MVTec-AD) and GeneralAD (the SOTA single-class general AD method). In our experiments, Dinomaly acheived the second-best on MVTec-AD and GeneralAD acheived the second-best on CIFAR-10. First, Dinomaly employed 392×392 resolution images and GeneralAD used 518×518 resolution images, while our method used 392×392 images. To further evaluate the effect of image resolution on performance, we conducted ablation studies with a reduced input resolution of 224×224. The results are provided in \cref{Table:MVTec-AD_size256} and \cref{cifar10_size256}. Second, both Dinomaly and GeneralAD used DINOv2 as their backbone network. The proposed method considerably outperformed Dinomaly in semantic AD and GeneralAD in multi-class settings.

\begin{table}[h]
   \footnotesize
    \vspace{-4mm}
    \begin{minipage}{0.45\linewidth}
  \caption{\textbf{Anomaly detection results in image-level AUROC(\%) on MVTec-AD under the multi-class setting.} Input images are resized to 256 × 256 and center-cropped to 224 × 224.}
\begin{center}
 \resizebox{0.5\linewidth}{!}{
  \begin{tabular}{c|c|c}
    \toprule
    \multicolumn{2}{c|}{Category}                                                & Ours\\
    \midrule
    \multirow{10}{*}{\rotatebox{90}{Object}}                  & Bottle           & 100  \\
\multicolumn{1}{c|}{}                                         & Cable            & 99.5     \\
\multicolumn{1}{c|}{}                                         & Capsule          & 96.7    \\
\multicolumn{1}{c|}{}                                         & Hazelnut         & 99.9     \\
\multicolumn{1}{c|}{}                                         & Metal Nut        & 100     \\
\multicolumn{1}{c|}{}                                         & Pill             & 97.8    \\
\multicolumn{1}{c|}{}                                         & Screw            & 96.4    \\
\multicolumn{1}{c|}{}                                         & Toothbrush       & 100   \\
\multicolumn{1}{c|}{}                                         & Transistor       & 98.8   \\
\multicolumn{1}{c|}{}                                         & Zipper           & 100  \\ \midrule
\multicolumn{1}{c|}{\multirow{5}{*}{\rotatebox{90}{Texture}}} & Carpet           & 100  \\
\multicolumn{1}{c|}{}                                         & Grid             & 99.8   \\
\multicolumn{1}{c|}{}                                         & Leather          & 100   \\
\multicolumn{1}{c|}{}                                         & Tile             & 100   \\
\multicolumn{1}{c|}{}                                         & Wood             & 94.5   \\ \midrule
\multicolumn{1}{c|}{}                                         & Mean             & 99.2   \\ \bottomrule
  \end{tabular}}
  \end{center}
  \label{Table:MVTec-AD_size256}
      \end{minipage}
\hfill
\vspace{-5mm}
 \begin{minipage}{0.45\linewidth}
    \centering
    \caption{\textbf{Semantic anomaly detection results in image-level AUROC(\%) on CIFAR-10 under the multi-class setting.} Normal class indices are shown in the left column, the remaining classes are considered as anomalies. Input images are resized and center-cropped to 224 × 224.}
    \label{cifar10_size256}
    \begin{center}
    \resizebox{0.6\linewidth}{!}{
    \begin{tabular}{c|c}
    \hline
    Normal Indices & Ours \\ \hline
    \{01234\}      &  96.7   \\
    \{56789\}      &  97.8   \\
    \{02468\}      &  98.5   \\
    \{13579\}      &  98.6   \\ \hline
    Mean           &  97.9   \\ \hline
    \end{tabular}}
    \end{center}
        \end{minipage}
        \vspace{-4mm}
\end{table}

\section{Histograms of Anomaly Scores for Testing Samples}
\begin{figure}[h]
\vspace{-3mm}
\begin{center}
\includegraphics[width=0.7\linewidth]{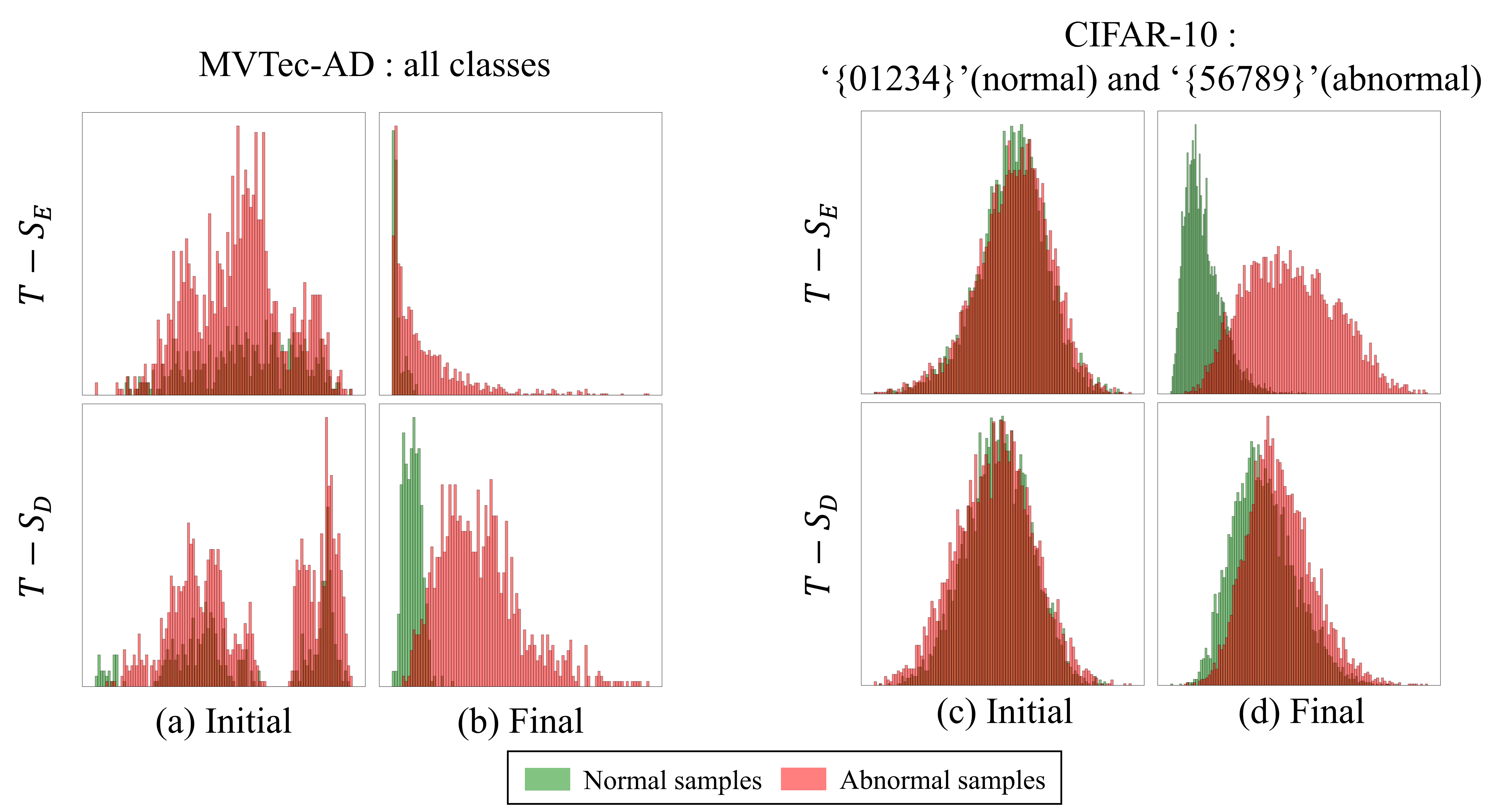}
\end{center}
\vspace{-4mm}
\caption{\textbf{Histograms of anomaly scores for testing samples.} 
}
\label{fig:feature_dist}
\vspace{-3mm}
\end{figure}

In the initial step of learning, both $T$-$S_{E}$ and $T$-$S_{D}$ fail to distinguish between normal and anomalies (\cref{fig:feature_dist}(a)). In final step, whereas the encoder still exhibits a significant overlap between the two, the decoder dichotomises normal from abnormal samples on MVTec-AD (\cref{fig:feature_dist}(b)). Before learning, both the student models do not separate normal from abnormal (\cref{fig:feature_dist}(c)). As learning proceeds, the encoder student well distinguishes the two classes on CIFAR-10 (\cref{fig:feature_dist}(d)).

\section{Additional Class-wise Results of Multi-class Anomaly Detection}

In \Cref{tab:MVTec-AD,tab:cifar10,tab:cifar100_multi-class_table,tab:fmnist_multi-class_table,tab:view_multi-class_table,tab:visa_multi-class_results,tab:loco_multi-class_results,tab:realiad_multi-class_results}, we present detailed class-wise results from our comprehensive multi-class anomaly detection experiments for MVTec-AD, CIFAR-10, VisA, Real-IAD, MVTec-LOCO, FMNIST, CIFAR-100 and View datasets. Additionally, for some methods where multi-class industrial AD accuracies were not explicitly reported in their original works, we supplemented these with referenced results from relevant studies \cite{dinomaly, ader}, ensuring a thorough comparison.

\begin{table}[hbt!]
    \vspace{-3mm}
  \caption{\textbf{Anomaly detection results in image-level AUROC(\%) on MVTec-AD under multi-class setting.}}
\begin{center}
  \tabcolsep=0.08cm
  {\footnotesize
  \resizebox{1\textwidth}{!}{
  \begin{tabular}{c|c|ccccccc|c}
    \toprule
    \multicolumn{2}{c|}{Category}                                     &\shortstack{RD4AD\\\cite{rd4ad}} &\shortstack{SimpleNet\\\cite{simplenet}} &\shortstack{DeSTSeg\\\cite{destseg}} & \shortstack{ReContrast\\\cite{recontrast}}  &\shortstack{UniAD\\\cite{uniad}}  &\shortstack{Dinomaly\\\cite{dinomaly}} & \shortstack{GeneralAD\\\cite{generalad}} & Ours\\
    \midrule
    \multirow{10}{*}{\rotatebox{90}{Object}}                  & Bottle           & 99.6                 & \textbf{100}                  & 98.7                & \textbf{100}                         & 99.7              & \textbf{100}                    & 25.8                        & \textbf{100}     \\
\multicolumn{1}{c|}{}                                         & Cable            & 84.1                 & 97.5                 & 89.5                & 95.6                        & 95.2              & \textbf{100}                    & 56.8                        & \underline{99.7}     \\
\multicolumn{1}{c|}{}                                         & Capsule          & 94.1                 & 86.9                 & 82.8                & 97.3                        & 86.9              & \underline{97.9}                   & 38.1                        & \textbf{98.5}    \\
\multicolumn{1}{c|}{}                                         & Hazelnut         & 60.8                 & 99.8                 & 98.9                & \textbf{100}                         & 99.8              & \textbf{100}                    & 51.9                        & \textbf{100}     \\
\multicolumn{1}{c|}{}                                         & Metal Nut        & 100                  & 99.2                 & 92.9                & \textbf{100}                         & 99.2              & \textbf{100}                    & 44.1                        & \textbf{100}     \\
\multicolumn{1}{c|}{}                                         & Pill             & 97.5                 & 93.7                 & 77.1                & 96.3                        & 93.7              & \underline{99.1}                   & 71.9                        & \textbf{99.3}    \\
\multicolumn{1}{c|}{}                                         & Screw            & 97.7                 & 76.7                 & 69.9                & 97.2                        & 87.5              & \underline{98.4}                   & 53.6                        & \textbf{98.7}    \\
\multicolumn{1}{c|}{}                                         & Toothbrush       & 97.2                 & 89.7                 & 71.7                & 96.7                        & 94.2              & \textbf{100}                    & 53.6                        & \textbf{100}     \\
\multicolumn{1}{c|}{}                                         & Transistor       & 94.2                 & \underline{99.2}                 & 78.2                & 94.5                        & \textbf{99.8}              & 99.0                   & 68.0                        & 99.0    \\
\multicolumn{1}{c|}{}                                         & Zipper           & 99.8                 & 95.8                 & 88.4                & 99.4                        & 95.8              & \textbf{100}                    & 77.0                        & \textbf{100}     \\ \midrule
\multicolumn{1}{c|}{\multirow{5}{*}{\rotatebox{90}{Texture}}} & Carpet           & 98.5                 & \textbf{99.8}     & 95.9                    & 98.3           & \textbf{99.8}         & \textbf{99.8}             & 50.2           & \textbf{99.8}      \\
\multicolumn{1}{c|}{}                                         & Grid             & 98.0                 & 98.2                 & 97.9                & 98.9                        & 98.2              & \textbf{99.9}                   & 59.9                        & \underline{99.8}    \\
\multicolumn{1}{c|}{}                                         & Leather          & \textbf{100}           & \textbf{100}        & 99.2           & \textbf{100}                    & \textbf{100}           & \textbf{100}             & 68.6                & \textbf{100}     \\
\multicolumn{1}{c|}{}                                         & Tile             & 98.3                 & 99.3                 & 97.0                & 99.5                        & 99.3              & \textbf{100}                    & 92.2                        & \textbf{100}     \\
\multicolumn{1}{c|}{}                                         & Wood             & 99.2                 & 98.6                 & \underline{99.9}                & 99.7                        & 98.6              & 99.8                   & 57.6                        & \textbf{100}     \\ \midrule
\multicolumn{1}{c|}{}                                         & Mean             & 94.6                 & 96.5                 & 89.2                & 98.2                        & 96.5              & \underline{99.6}                   & 56.8                        & \textbf{99.7}    \\ \bottomrule
  \end{tabular}}}
  \end{center}
  \label{tab:MVTec-AD}
\vspace{-5mm}
\end{table}

\begin{table}[hbt!]
\setlength{\tabcolsep}{1pt}
    \footnotesize
    \centering
    \vspace{-3mm}
    \caption{\textbf{Semantic anomaly detection results in AUROC(\%) on CIFAR-10 under multi-class setting.} Normal classes indices are shown in the leftmost column, the remaining classes are considered as anomalies.}
    \label{tab:cifar10}
    \begin{center}
  \resizebox{0.8\textwidth}{!}   
    {\begin{tabular}{c|cccccc|c}
    \hline
    \multirow{2}{*}{\shortstack{Normal\\Indices}} & \multirow{2}{*}{\shortstack{Transformaly\\\cite{transformaly}}} &\multirow{2}{*}{\shortstack{MSC\\ \cite{msc}}}& \multirow{2}{*}{\shortstack{UniAD\\ \cite{uniad}}} & \multirow{2}{*}{\shortstack{MINT-AD\\\cite{mint-ad}}} & \multirow{2}{*}{\shortstack{GeneralAD\\\cite{generalad}}} & \multirow{2}{*}{\shortstack{Dinomaly\\\cite{dinomaly}}} & \multirow{2}{*}{Ours} \\ 
    & & & & & & & \\ \hline
    \{01234\}      & 92.3                            &  84.5       & 84.4  &  86.7   & \underline{92.7}   & 63.7  &  \textbf{96.6}    \\
    \{56789\}      & \underline{90.0}                            &  88.4       & 80.9  &  83.3   & 88.2   & 64.8  &  \textbf{97.6}    \\
    \{02468\}      & 97.4                            &  92.6       & 93.0  &  93.6   & \underline{98.3}   & 78.8  &  \textbf{98.5}    \\
    \{13579\}      & 94.5                            &  93.2       & 90.6  &  92.6   & \underline{96.2}   & 78.2  &  \textbf{98.4}    \\ \hline
    mean           & 93.6                            &  89.7       & 87.2  &  89.1   & \underline{93.9}   & 71.4  &  \textbf{97.8}    \\ \hline
    \end{tabular}%
    }
    \end{center}
    \vspace{-5mm}
\end{table}

\begin{table}[hbt!]
    \centering
    \vspace{-2mm}
    \caption{\textbf{Anomaly detection results in AUROC(\%) on VisA under multi-class setting.}}
    \vspace{-1mm}
    \label{tab:visa_multi-class_results}
    \begin{center}
    \resizebox{1\textwidth}{!}{
    \begin{tabular}{c|cccccccc|c}
    \hline
    Category    &KNN\cite{knn}  &Transformaly\cite{transformaly}  &RD4AD\cite{rd4ad} &SimpleNet\cite{simplenet} & DeSTSeg\cite{destseg} &UniAD\cite{uniad} &Dinomaly\cite{dinomaly}&GeneralAD\cite{generalad}                 & Ours \\ 
    \hline
    candle      &71.5  &61.2  &92.3   &84.1   &94.9   &94.1   &\textbf{98.7}   & 91.3 & \textbf{98.7}\\
    capsules    &61.9  &70.4  &82.2   &74.1   &87.1   &55.6   &\underline{98.6} & 86.3 & \textbf{98.7}\\
    cashew      &95.8  &89.6  &92.0   &88.0   &92.0   &92.8   &\textbf{98.7}   & 85.9 & \underline{98.4}\\
    chewinggum  &92.8  &88.4  &94.9   &96.4   &95.8   &96.3   &\textbf{99.8}   & 94.3 &\textbf{99.8}\\
    fryum       &86.3  &38.9  &95.3   &88.4   &92.1   &83.0   &\underline{98.8} & 88.9 & \textbf{98.9}\\
    macaroni1   &74.6  &58.2  &75.9   &85.9   &76.6   &79.9   &\textbf{98.0}    &84.4 & \underline{97.9}\\
    macaroni2   &57.3  &43.9  &88.3   &68.3   &68.9   &71.6   &\underline{95.9}  & 82.9& \textbf{96.5}\\
    pcb1        &92.0  &73.7  &96.2   &91.6   &87.6   &92.8   &\underline{99.1}  & 96.1& \textbf{99.6}\\
    pcb2        &87.0  &69.4  &97.8   &92.4  &86.5    &87.8   &\underline{99.3}  & 91.2& \textbf{99.4}\\
    pcb3        &70.2  &66.8  &96.4   &89.1  &93.7    &78.6   &\underline{98.9}  & 92.4& \textbf{99.2}\\
    pcb4        &89.9  &52.5  &\textbf{99.9}   &97.0  &97.8    &98.8   &99.8  & 97.7& \textbf{99.9}\\
    pipe\_fryum &94.2  &66.9 &97.9  &90.8 &94.1    &94.7   &\textbf{99.2}  &91.2 & \underline{98.3}\\
    \hline
    mean        &81.1   &65.0  & 92.4 &87.2  &88.9    &88.8   &\underline{99.8}  & 90.2 & \textbf{99.9}\\                  
    \hline
    \end{tabular}%
    }
    \vspace{-5mm}
    \end{center}
\end{table}

\begin{table}[hbt!]
    \centering
    \vspace{-3mm}
    \caption{\textbf{Anomaly detection results in AUROC(\%) on Real-IAD under multi-class setting.}}
    \label{tab:realiad_multi-class_results}
    \begin{center}
    \resizebox{1\textwidth}{!}   
    {{\tiny
    \begin{tabular}{c|ccccc|c}
    \hline
Category            &RD4AD\cite{rd4ad} &SimpleNet\cite{simplenet} & DeSTSeg\cite{destseg} &UniAD\cite{uniad} &Dinomaly\cite{dinomaly}     & Ours \\ 
    \hline
audiojack	           & 76.2          & 58.4                     & 81.1                  & 81.4             & \textbf{86.8}              &\textbf{86.8}\\
bottle\_cap	           & 89.5          & 54.1                     & 78.1                  & \textbf{92.5}    & \underline{89.9}           & 87.2\\
button\_battery        & 73.3          & 52.5                     & 86.7                  & 75.9             & \underline{86.6}           &\textbf{87.0}\\
end\_cap	           & 79.8          & 51.6                     & 77.9                  & 80.9             & \textbf{87.0}              &\underline{84.6}\\
eraser	               & 90.0          & 46.4                     & 84.6                  & \underline{90.3}             & \underline{90.3}           &\textbf{92.9}\\
fire\_hood	           & 78.3          & 58.1                     & 81.7                  & 80.6             & \underline{83.8}                         &\textbf{84.3}\\
mint	               & 65.8          & 52.4                     & 58.4                  & 67.0             & \textbf{73.1}                     & \underline{72.8}\\
mounts	               & 88.6          & 58.7                     & 74.7                  & 87.6             & \textbf{90.4}                   & \underline{89.6}\\
pcb	                   & 79.5          & 54.5                     & 82.0                  & 81.0             & \textbf{92.0}                                 &\underline{91.9}\\
phone\_battery	       & 87.5          & 51.6                     & 83.3                  & 83.6             & \textbf{92.9}                                   & \underline{91.8}\\
plastic\_nut	       & 80.3          & 59.2                     & 83.1                  & 80.0             & \textbf{88.3}                                   &\underline{86.1}\\
plastic\_plug	       & 81.9          & 48.2                     & 71.7                  & 81.4             & \textbf{90.5}                                  & \underline{87.0}\\
porcelain\_doll	       & 86.3          & 66.3                     & 78.7                  & \textbf{85.1}             & \textbf{85.1}                                  &84.7\\
regulator	           & 66.9          & 50.5                     & 79.2                  & 56.9             & \underline{85.2}                                   &\textbf{86.7}\\
rolled\_strip\_base	   & 97.5          & 59.0                     & 96.5                  & 98.7             & \underline{99.2}                                   &\textbf{99.3}\\
sim\_card\_set	       & 91.6          & 63.1                     & 95.5                  & 89.7             & \underline{95.8}                                   &\textbf{96.0}\\
switch	               & 84.3          & 62.2                     & 90.1                  & 85.5             & \textbf{97.8}                                   &\underline{97.3}\\
tape	               & 96.0          & 49.9                     & 94.5                  & \textbf{97.2}     & \underline{96.9}                                   &96.2\\
terminalblock	       & 89.4          & 59.8                     & 83.1                  & 87.5             & \textbf{96.7}                                   & \underline{94.0}\\
toothbrush	           & 82.0          & 65.9                     & 83.7                  & 78.4             & \textbf{90.4}                                   & \underline{87.4}\\
toy	                   & 69.4          & 57.8                     & 70.3                  & 68.4             & \textbf{85.6}                                   & \underline{85.4}\\
toy\_brick	           & 63.6          & 58.3                     & 73.2                  & \textbf{77.0}             & 72.3                                   & \underline{73.0}\\
transistor1	           & 91.0          & 62.2                     & 90.2                  & 93.7             & \textbf{97.4}                                   &\textbf{97.4}\\
u\_block	           & 89.5          & 62.4                     & 80.1                  & 88.8             & \underline{89.9}                                   &\textbf{90.6}\\
usb	                   & 84.9          & 57.0                     & 87.8                  & 78.7             & \textbf{92.0}                                   &\underline{91.2}\\
usb\_adaptor	       & 71.1          & 47.5                     & 80.1                  & 76.8             & \underline{81.5}                                   &\textbf{83.5}\\
vcpill	               & 85.1          & 59.0                     & 83.8                  & 87.1             & \textbf{92.0}                                     &\underline{91.0}\\
wooden\_beads	       & 81.2          & 55.1                     & 82.4                  & 78.4             & \underline{87.3}                                  &\textbf{88.1}\\
woodstick	           & 76.9          & 58.2                     & 80.4                  & \underline{80.8}    & \textbf{84.0}                                   &79.7\\
zipper	               & 95.3          & 77.2                     & 96.9                  & 98.2             & \textbf{99.1}                                     &\underline{98.8}\\
    \hline
mean                   & 82.4          & 57.2                     & 82.3                  & 83.0             & \textbf{89.3}                                    & \underline{88.7}  \\     
    \hline
    \end{tabular}%
    }}
    \end{center}
    \vspace{-5mm}
\end{table}

\begin{table}[hbt!]
    \centering
    \vspace{-3mm}
    \caption{\textbf{Anomaly detection results in AUROC(\%) on MVTec-LOCO under multi-class setting.}}
    \label{tab:loco_multi-class_results}
    \begin{center}
    \resizebox{1\textwidth}{!}   
    {\begin{tabular}{c|cccccccc|c}
    \hline
    Category             &KNN\cite{knn}  &Transformaly\cite{transformaly}&RD4AD\cite{rd4ad} &SimpleNet\cite{simplenet} & DeSTSeg\cite{destseg} &UniAD\cite{uniad} &Dinomaly\cite{dinomaly} &GeneralAD\cite{generalad}   & Ours \\ 
    \hline
    breakfast\_box       &73.1          & 65.1                          & 63.4              & 89.4                     &84.0                  & 77.5              &\underline{91.6}           &  90.9          & \textbf{93.5} \\
    juice\_bottle        &88.6          & 59.4                          & 89.5              & 96.0                     &\textbf{99.5}                  & 95.4               &91.2              &    92.2     & \underline{96.4} \\
    pushpins             &64.4          & 48.9                          & 71.5              & 76.5                     &\textbf{72.2}                  & \underline{71.3}               &69.1      &      68.4           & 70.9 \\
    screw\_bag           &\underline{72.3}          & 54.8                          & 63.7              & 65.3                     &70.0                  & 65.0               &\underline{72.3}    &       69.5            & \textbf{77.0} \\
    splicing\_connectors &75.5          & 57.9                          & 80.5              & 81.7                     &80.1                  & 84.2               &\underline{88.4}                &   \underline{88.4}    & \textbf{92.7} \\
    \hline
    mean                 &74.8          & 57.2                          &73.7               &81.8                       &81.2                 & 78.7               &\underline{82.5}              &    81.9     & \textbf{86.1}  \\                  
    \hline
    \end{tabular}%
    }
    \end{center}
        \vspace{-5mm}
\end{table}

\begin{table}[hbt!]
    \small
    \centering
    \vspace{-1mm}
    \caption{\textbf{Semantic anomaly detection results in AUROC(\%) on FMNIST under multi-class setting.} Normal classes indices are shown in the leftmost column, the remaining classes are considered as anomalies.}
    \vspace{-1mm}
    \label{tab:fmnist_multi-class_table}
    \begin{center}
  \resizebox{1\textwidth}{!}   
    {\begin{tabular}{c|ccccc|c}
    \hline
    Normal Indices & KNN\cite{knn}&Transformaly\cite{transformaly} &MSC\cite{msc}&   Dinomaly\cite{dinomaly}&GeneralAD\cite{generalad} & Ours \\ \hline
    \{01234\}      & 92.2                            & 92.3        &\textbf{94.3}   & \underline{93.8}    &92.5    & 93.2        \\
    \{56789\}      & 75.8                            & 76.0        &75.7   & 69.0    &\textbf{88.0}    & \underline{78.9}        \\
    \{02468\}      & \textbf{95.1}                            & 81.7        &93.8   & 72.0    &\underline{94.4}    & 91.8        \\
    \{13579\}      & 89.7                            & \underline{93.8}        &93.5   & \textbf{95.1}    &94.5    & 93.4        \\ \hline
    mean           & 88.2                            & 85.6        &\underline{89.3}   &82.5     &\textbf{92.4}    &88.3       \\ \hline
    \end{tabular}%
    }
    \end{center}
    \vspace{-6mm}
\end{table}

\begin{table}[hbt!]
    \small
    \centering
    \vspace{-2mm}
    \caption{\textbf{Semantic anomaly detection results in AUROC(\%) on CIFAR-100 under multi-class setting.} Normal classes indices are shown in the leftmost column, the remaining classes are considered as anomalies.}
    \label{tab:cifar100_multi-class_table}
    \begin{center}
  \resizebox{1\textwidth}{!}   
    {\begin{tabular}{c|ccccc|c}
    \hline
    Normal Indices                                      & KNN\cite{knn}&Transformaly\cite{transformaly} &MSC\cite{msc}&   Dinomaly\cite{dinomaly}&GeneralAD\cite{generalad} & Ours \\ \hline
    \{0, 1, 2, 3, 4, 5, 6, 7, 8, 9\}                 &  85.3                           & 81.3         &81.1   &65.2       &\underline{89.3}    & \textbf{90.2}        \\
    \{10, 11, 12, 13, 14, 15, 16, 17, 18, 19\}      &   \underline{89.4}                          & \underline{89.4}         &88.4  & 67.8       &87.9    & \textbf{95.4}       \\
    \{0, 2, 4, 6, 8, 10, 12, 14, 16, 18\}           &   87.9                          & 86.5        &83.2   & 64.2      &\underline{91.4}    & \textbf{92.3}        \\
    \{1, 3, 5, 7, 9, 11, 13, 15, 17, 19\}           &   85.6                          & 83.4        &81.9   & 63.4      &\underline{90.1}    & \textbf{91.4}        \\ \hline
    mean                                            & 87.0                            &85.2        &83.7   &65.1     &\underline{89.7}    & \textbf{92.3}       \\ \hline
    \end{tabular}%
    }
    \end{center}
    \vspace{-5mm}
\end{table}

\begin{table}[hbt!]
    \normalsize
    \centering
    \vspace{-1mm}
    \caption{\textbf{Semantic anomaly detection results in AUROC(\%) on View under multi-class setting.} Normal classes are shown in the leftmost column, the remaining classes are considered as anomalies.}
    \label{tab:view_multi-class_table}
    \begin{center}
  \resizebox{1\textwidth}{!}   
    {\begin{tabular}{c|ccccc|c}
    \hline
    Normal Classes  & KNN\cite{knn}&Transformaly\cite{transformaly} &MSC\cite{msc}&   Dinomaly\cite{dinomaly}&GeneralAD\cite{generalad} & Ours \\ \hline
    \{buildings, forest, glacier\}      & 49.5                &\underline{86.4}         &66.9   & 49.4    &69.8    & \textbf{88.3}        \\
    \{mountain, sea, street\}      & 71.8                    & \underline{81.2}        &65.6    & 78.9    &71.0    & \textbf{85.8}        \\
    \{buildings, glacier, sea\}      & 69.7                  & 71.8        &64.0   & \underline{71.9}   &69.7    &  \textbf{85.7}       \\
    \{forest, mountain, street\}      & 70.8                 & \underline{82.9}        &69.8   & 66.7    &71.3    &  \textbf{89.5}       \\ \hline
    mean           & 65.6                            &\underline{82.2}        &66.6   &66.6    &70.5    &\textbf{87.3}       \\ \hline
    \end{tabular}%
    }
    \end{center}
    \vspace{-5mm}
\end{table}

\section{Multi-class Anomaly Detection Results with Additional Metrics}

In \Cref{tab:pixel_level_comparison}, we present anomaly detection results with various metrics from our comprehensive multi-class anomaly detection experiments for MVTec-AD and VisA datasets. Additionally, for some methods where multi-class industrial AD accuracies were not explicitly reported in their original works, we supplemented these with referenced results from relevant studies \cite{dinomaly, ader}, ensuring a thorough comparison. Note our method is superior to best specialist models in most metrics, while the specialist models to industrial AD do not generalise to semantic AD tasks. 

\begin{table}[hbt!]
  \centering
  \tiny
  \caption{\textbf{Multi-class anomaly detection results with additional metrics.}}
   \resizebox{0.9\textwidth}{!}{
    \begin{tabular}{ccccccccc}
    \toprule
    \multirow{2}[2]{*}{Dateset} & \multirow{2}[2]{*}{Method} & \multicolumn{3}{c}{Image-level} & \multicolumn{4}{c}{Pixel-level} \\
\cmidrule(r){3-5} \cmidrule(l){6-9} 
& &\multicolumn{1}{c}{AUROC} & \multicolumn{1}{c}{AP} & \multicolumn{1}{c}{$F_1$-max} & \multicolumn{1}{c}{AUROC} & \multicolumn{1}{c}{AP} & \multicolumn{1}{c}{$F_1$-max} & \multicolumn{1}{c}{AUPRO} \\
\hline
    \multirow{10}[0]{*}{\textbf{MVTec-AD}~\cite{mvtecad}}
          & DRAEM~\cite{dream} & 54.5 & 76.3 & 83.6 & 47.6 & 3.2  & 6.7 & 14.3 \\
          & RD4AD~\cite{rd4ad} & 94.6  & 96.5  & 95.2  & 96.1  & 48.6  & 53.8  & 91.1 \\
          & RealNet~\cite{realnet} & 84.8 & 94.1 & 90.9 & 72.6 & 48.2 & 41.4 & 56.8 \\
          & SimpleNet~\cite{simplenet} & 95.3 & 98.4  & 95.8  & 96.9  & 45.9  & 49.7  & 86.5\\
          & DeSTSeg~\cite{destseg} & 89.2  & 95.5  & 91.6  & 93.1  & 54.3  & 50.9  & 64.8\\
          & UniAD~\cite{uniad} & 96.5  & 98.8  & 96.2  & 96.8  & 43.4  & 49.5  & 90.7\\
          & ReContrast~\cite{recontrast} & 98.3 & 99.4 & 97.6 & 97.1 & 60.2 & 61.5 & 93.2 \\
          & RD++~\cite{rd4ad++} & 97.9 & 98.8 & 96.4 & 97.3 & 54.7 & 58.0 & 93.2 \\
          & Dinomaly~\cite{dinomaly} &\underline{99.6} & \textbf{99.8} & \underline{99.0} & \textbf{98.4} & \textbf{69.3} & \textbf{69.2} & \textbf{94.8} \\
          & \textbf{Ours} &\textbf{99.7} & \textbf{99.8} & \textbf{99.1} & \textbf{98.4} & \underline{69.0} & \underline{69.0} & \underline{94.6} \\

    \hline
    \multirow{10}[0]{*}{\textbf{VisA}~\cite{visa}} 
          & DRAEM~\cite{dream} & 55.1 & 62.4 & 72.9 & 37.5 & 0.6 & 1.7 & 10.0 \\
          & RD4AD~\cite{rd4ad} & 92.4  & 92.4  & 89.6 & 98.1 & 38.0  & 42.6  &  91.8 \\
          & RealNet~\cite{realnet} & 71.4 & 79.5 & 74.7 & 61.0 & 25.7 & 22.6 & 27.4 \\
          & SimpleNet~\cite{simplenet} & 87.2  & 87.0  & 81.8  & 96.8  & 34.7  & 37.8  & 81.4\\
          & DeSTSeg~\cite{destseg} & 88.9  & 89.0  & 85.2  & 96.1  & 39.6 & 43.4  & 67.4\\
          & UniAD~\cite{uniad} & 88.8  & 90.8  & 85.8  & 98.3  & 33.7  & 39.0  & 85.5\\
          & ReContrast~\cite{recontrast} & 95.5 & 96.4 & 92.0 & 98.5 & 47.9 & 50.6 & 91.9  \\
          & RD++~\cite{rd4ad++} & 93.9 & 94.7 & 90.2 & 98.4 & 42.3 & 46.3 & 91.9 \\
          & Dinomaly~\cite{dinomaly} & \textbf{98.7} & \textbf{98.9} & \underline{96.2} & \underline{98.7} & \underline{53.2} & \textbf{55.7} & \textbf{94.5} \\
          & \textbf{Ours} & \textbf{98.7} & \textbf{98.9} & \textbf{96.3} & \textbf{98.8} & \textbf{53.4} & \underline{55.6} & \textbf{94.5} \\
    \bottomrule
    \end{tabular}%
    }
  \label{tab:pixel_level_comparison}%
\end{table}%

\section{Detailed Comparison with DSKD}

\begin{figure}[ht!]
        \begin{minipage}{0.48\linewidth}
        \begin{subfigure}{\linewidth}
        \centering
        \includegraphics[width=0.8\linewidth]{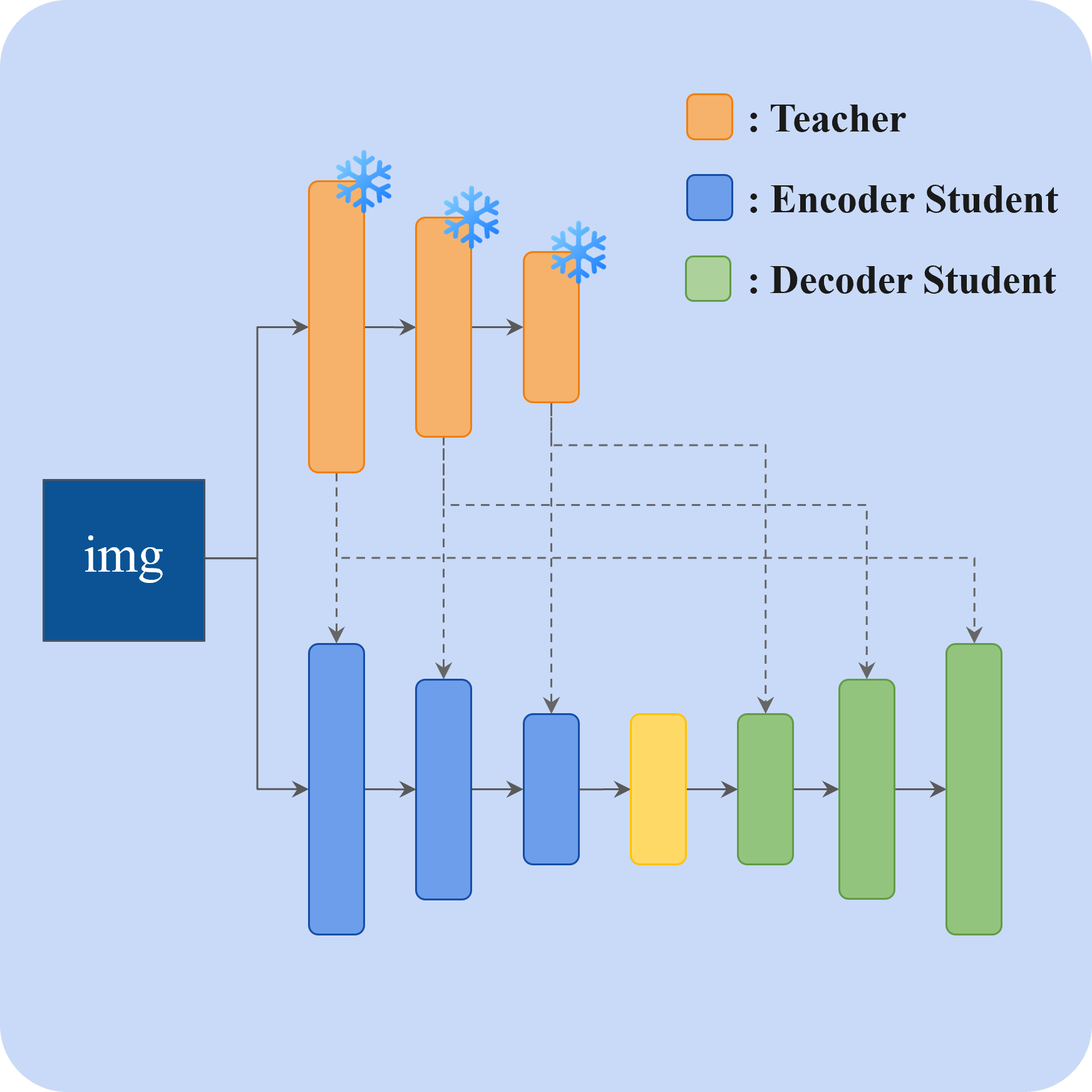}
        \vspace{-2mm}
        \caption{DSKD}
        \label{fig:DSKD}
       \end{subfigure}
       \end{minipage}
       \hfill
       \begin{minipage}{0.48\linewidth}
       \begin{subfigure}{\linewidth}
       \centering
        \includegraphics[width=0.8\linewidth]{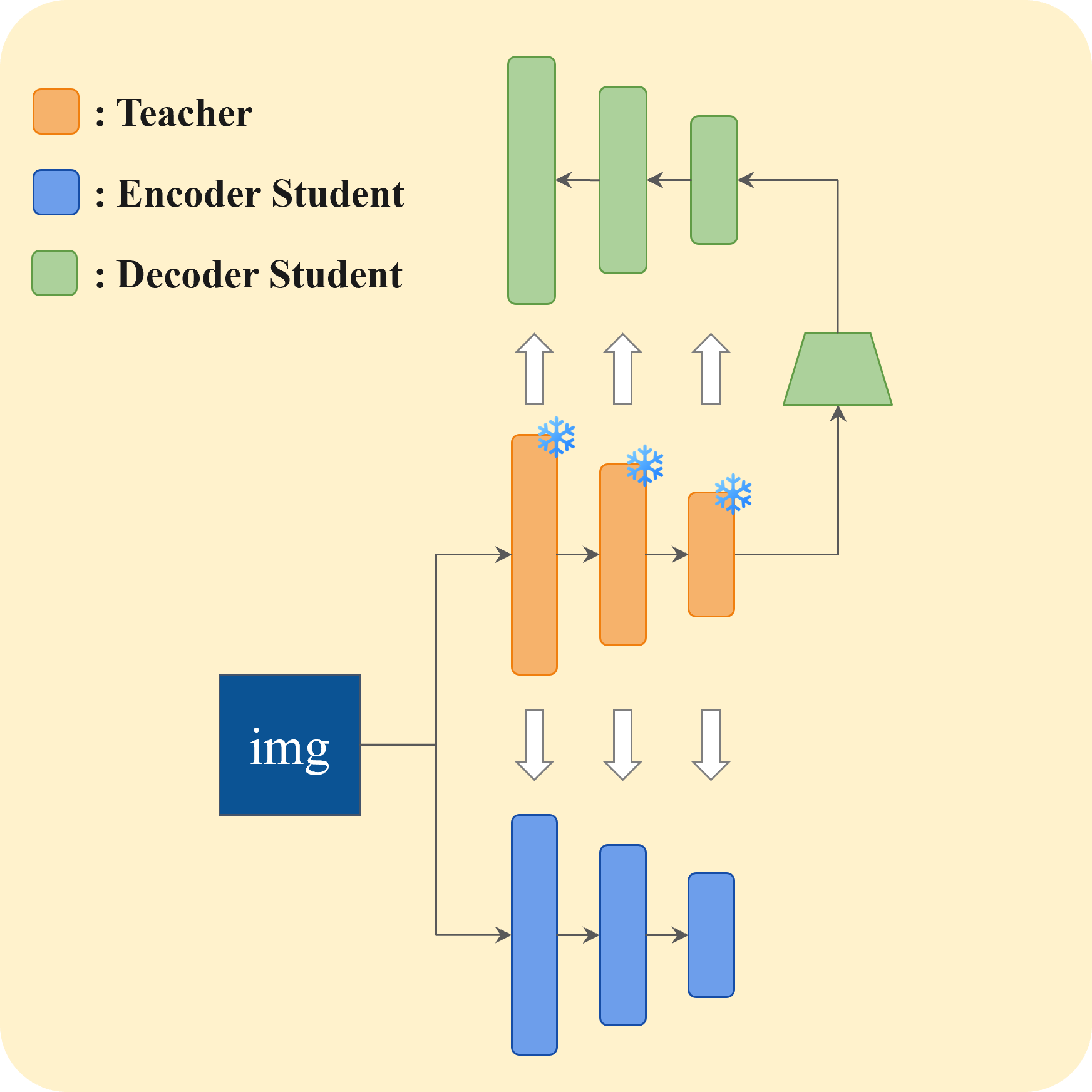}
        \vspace{-2mm}
        \caption{Our method}
        \label{fig:ours}
       \end{subfigure}
       \end{minipage}
    \vspace{-2mm}
    \caption{\textbf{Architectures of dual-student KD methods} (a) DSKD \cite{dualstudent}, (b) Our method}
    \label{fig:comparison_dskd}
    \vspace{-4mm}
\end{figure}

Although DSKD \cite{dualstudent} adopts a dual-student concept similar to our method, the two approaches diverge significantly in that our model is specifically designed with semantic anomaly detection in mind. The main differences between our method and DSKD are as follows:

(1) DSKD and our method are architecturally distinct, as illustrated in \Cref{fig:comparison_dskd}. 
DSKD utilizes two student networks, connected through a bottleneck block, which learns from a shared teacher network in the encoding and decoding stages, respectively. In contrast, our method connects the teacher network and decoder via a bottleneck, while the teacher and encoder pair exists independently to capture semantic representations.

(2) DSKD employs features extracted from ResNet18 in both the student encoder and student decoder. Our method is based on a Vision Transformer (ViT) and deliberately utilizes different types of representations for each branch. Specifically, the encoder-decoder branch utilizes patch features, while the encoder-encoder branch leverages class tokens. Since the class token encapsulates global representations of entire images, it is more effective than patch tokens in the encoder-encoder branch, which is tailored for semantic anomaly detection. This is empirically validated by the results shown in \Cref{tab:token_ablation}.

(3) Another key difference lies in the inference strategies of DSKD and our method. DSKD computes anomaly maps by combining the Euclidean 
 distance and cosine similarity between the teacher and decoder features. While this approach captures local-level discrepancies, it is not well-suited for semantic anomaly detection, which requires global contextual understanding. In contrast, our method extracts two complementary types of anomaly scores. A local anomaly score is obtained using cosine similarity between teacher and decoder features, which is particularly effective for industrial anomaly detection. A semantic anomaly score is computed using the mean squared error (MSE) between the class tokens of the teacher and encoder, capturing high-level semantic deviations crucial for semantic anomaly detection. To integrate these two perspectives, we adopt a Noisy-OR ensemble mechanism, which effectively fuses both local and global anomaly cues. This design enables our method to perform robustly across both industrial and semantic anomaly detection scenarios.

Consequently, DSKD shows limited performance in semantic anomaly detection. A detailed performance comparison with our method is presented in \Cref{tab:cifar10_DSKD_vs_ours}.

\begin{table}[hbt!]
    \small
    \centering
    \caption{\textbf{Semantic anomaly detection performance comparison between DSKD and our method on CIFAR-10 under the multi-class setting.} Normal classes indices are shown in the leftmost column, the remaining classes are considered as anomalies.}
    \label{tab:cifar10_DSKD_vs_ours}
    \begin{center}
  \resizebox{0.4\textwidth}{!}   
    {\begin{tabular}{c|c|c}
    \hline
    Normal Indices              & DSKD \cite{dualstudent}       & Ours \\ \hline
    \{0, 1, 2, 3, 4\}           & 56.7                          & \textbf{96.6}        \\
    \{5, 6, 7, 8, 9\}           & 46.8                          & \textbf{97.6}       \\
    \{0, 2, 4, 6, 8\}           & 59.7                          & \textbf{98.5}        \\
    \{1, 3, 5, 7, 9\}           & 46.3                          & \textbf{98.4}        \\ \hline
    mean                        & 52.1                          & \textbf{97.8}       \\ \hline
    \end{tabular}%
    }
    \end{center}
   \vspace{-2mm}
\end{table}

\end{document}